\pdfoutput=1
\documentclass[11pt]{article}

\usepackage[preprint]{acl}

\usepackage{times}
\usepackage{latexsym}
\usepackage{enumitem}
\usepackage{booktabs}
\usepackage{tabularx}
\usepackage[T1]{fontenc}
\usepackage{CJKutf8}
\usepackage{hyperref}
\usepackage[utf8]{inputenc}

\usepackage{microtype}

\usepackage{inconsolata}

\usepackage{graphicx}

\usepackage{amssymb}
\usepackage{amsmath}
\usepackage{multirow}
\usepackage{booktabs}
\usepackage{afterpage}
\usepackage{titlesec} 
\usepackage{ragged2e}
\usepackage{graphicx} % 引入 graphicx 包
\usepackage{array}
\usepackage{dsfont}
\usepackage[linesnumbered,ruled,vlined]{algorithm2e}
\SetKwComment{Comment}{$\triangleright$\ }{}
\SetKwInput{KwInit}{Initialize}

\title{Towards Meta-Cognitive Knowledge Editing for Multimodal LLMs}

\author{%
  \textbf{Zhaoyu Fan}$^1$, \textbf{Kaihang Pan}$^{1,\dagger}$, \textbf{Mingze Zhou}$^1$, 
  \textbf{Bosheng Qin}$^1$, \textbf{Juncheng Li}$^{1,\ddag}$, \\ 
  \textbf{Shengyu Zhang}$^1$,
  \textbf{Wenqiao Zhang}$^1$, \textbf{Siliang Tang}$^1$, \textbf{Fei Wu}$^1$, \textbf{Yueting Zhuang}$^1$\\ \\
  Zhejiang University$^1$ \\
  \texttt{\{zyfan, kaihangpan, mingze, bsqin, junchengli\}@zju.edu.cn} \\
  \texttt{\{sy\_zhang, wenqiaozhang, siliang, wufei, yzhuang\}@zju.edu.cn} \\
  \vspace{-10pt}
}

\setlist[itemize]{itemsep=0pt, parsep=0pt, topsep=0pt}
\begin{document}
\begin{CJK*}{UTF8}{gbsn}
\maketitle
\renewcommand{\thefootnote}{\fnsymbol{footnote}}
% \footnotetext[1]{~Equal Contribution.}
\footnotetext[2]{~Project Leader.}
\footnotetext[3]{~Corresponding Author.}
\renewcommand{\thefootnote}{\arabic{footnote}}

\begin{abstract}
Knowledge editing enables multimodal large language models (MLLMs) to efficiently update outdated or incorrect information. However, existing benchmarks primarily emphasize cognitive-level modifications while lacking a focus on deeper meta-cognitive processes. To bridge this gap, we introduce \textbf{CogEdit}, a novel benchmark designed to evaluate MLLMs’ meta-cognitive knowledge editing abilities across three levels: (1) \textbf{Counterfactual-Driven Editing}, assessing self-awareness of knowledge correctness changes; (2) \textbf{Boundary Constraint Editing}, ensuring appropriate generalization without unintended interference; and (3) \textbf{Noise-Robust Editing}, promoting reflective evaluation of uncertain information. To advance meta-cognitive editing, we propose \textbf{MIND} (\textbf{M}eta-cognitive \textbf{IN}tegrated \textbf{D}ynamic Knowledge Editing), a framework that constructs a meta-knowledge memory for self-awareness, employs game-theoretic interactions to monitor knowledge activation, and incorporates label refinement for noise-robust updates. Extensive experiments show that \textbf{MIND} significantly outperforms existing cognitive editing approaches, achieving strong performance on both traditional and meta-cognitive knowledge editing benchmarks.

\end{abstract}

\section{Introduction}

Knowledge Editing~\cite{huang2023transformer, de2021editing, mitchell2021fast, meng2022locating, li2024pmet, zheng2023can, mitchell2022memory, zhong2023mquake, wang2024wiserethinkingknowledgememory, chen2024lifelong, wang2024lemoe, meng2022mass} offers an efficient way for updating large language models (LLMs) to correct errors or outdated information without negative side effects, which has been further extended to the editing of multimodal large language models (MLLMs), known as multimodal editing.
Previous research suggests that the process of knowledge editing closely resembles \textbf{human cognition}~\cite{pan2025towards,zhang2024comprehensive}. 
It enables MLLMs to incorporate new knowledge and rectify their previous misconceptions.
Three common metrics~\cite{DBLP:conf/emnlp/0008TL0WC023} are used to evaluate the effectiveness of this cognitive process: \textbf{Reliability} (accepting the new knowledge to achieve desired editing outcome); \textbf{Generality} (transfering the newly acquired knowledge to similar scenarios); \textbf{Locality} (unrelated knowledge remaining unchanged without mistakenly altered).

\begin{figure}[t]
    \centering
    % \hspace*{-15pt}
    \includegraphics[width=0.45\textwidth]{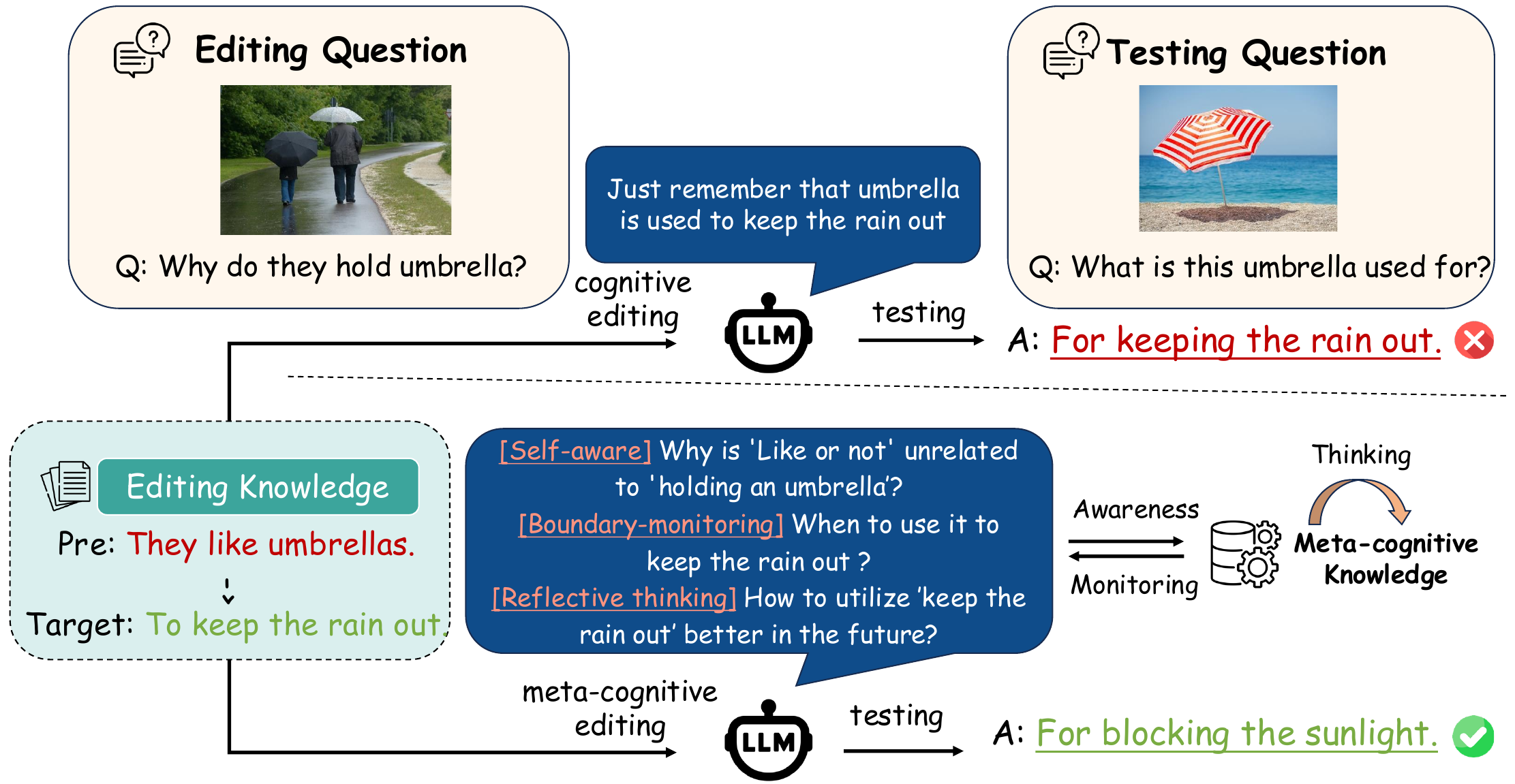}
    \vspace{-8pt}
    \caption{Difference between Cognitive Editing and Meta-cognitive Editing. Meta-cognitive editing requires ``thinking about thinking'', additionally involving self-awareness, boundary monitoring, reflective thinking.}
    \label{fig:intro_diff}
    \vspace{-15pt}
\end{figure}

While cognition is vital for learning and decision-making, it is insufficient alone. Firstly, cognitive editing simply identifies outdated knowledge and replaces it with new information mechanically, without recognizing when the old knowledge is no longer usable. Moreover, it recognizes knowledge only at a surface level, without truly understanding why the new information is correct. Additionally, it lacks reflective thinking about the cognitive process itself—crucial for refining updates, especially in the presence of noisy inputs.

The difference between cognitive and meta-cognitive editing is shown in Figure~\ref {fig:intro_diff}, to address the limitations of cognitive editing, effective knowledge editing should be \textbf{\textsc{meta-cognitive}}\cite{georghiades2004general, lai2011metacognition, veenman2006metacognition}. Meta-cognition encompasses a model's ability to assess and modify its own reasoning and learning processes. In contrast to basic cognition, meta-cognition emphasizes the capacity to ``think about thinking'' and incorporates three additional levels.

\begin{itemize}[leftmargin=*]
    \item \textbf{Level 1: Self-Awareness}~\cite{blakemore2003self}.
    % The system identifies and updates outdated or incorrect knowledge but does not fully comprehend the reasoning behind the change or the broader implications of the new knowledge.
    The system, while replacing outdated knowledge with new knowledge, is capable of self-awareness regarding the reasons why the old knowledge was incorrect and needed to be replaced in the corresponding context.

    \item \textbf{Level 2: Boundary Monitoring}~\cite{li2024knowledge}.  
    % excessive generalized to seemingly similar but unrelated samples
    % The system oversees and controls the conditions under which new knowledge is utilized, ensuring appropriate application while avoiding misgeneralization across different contexts.
    The system is capable of monitoring the utilization boundary of new knowledge, preventing it from being excessively overgeneralized and thereby avoiding the erroneous overshadowing of truly relevant knowledge.

    \item \textbf{Level 3: Reflective Thinking}~\cite{rodgers2002defining}.  
    % The system evaluates its knowledge update processes, refines its adaptation strategies，and enhances knowledge integration through self-assessment and learning from past modifications.
    The system evaluates its knowledge update processes and refines its adaptation strategies. It assesses external inputs to determine whether they should be accepted and, if so, which portions are relevant for integration.

\end{itemize}

\begin{figure}[t]  % 't' 表示图片在页面顶部
    \centering
    % \hspace*{-15pt}
    \includegraphics[width=0.5\textwidth]{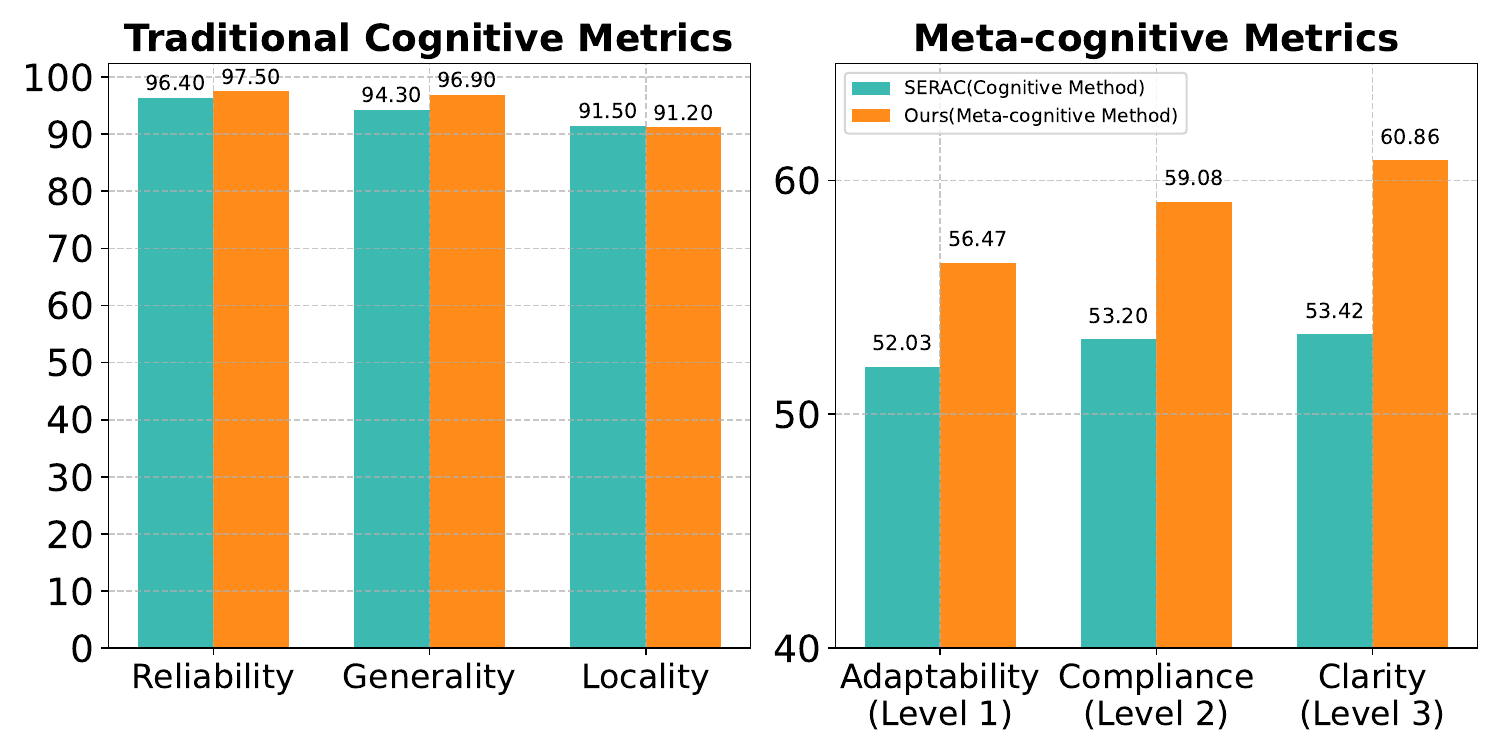}
    % \caption{Comparison between Cognitive and Meta-cognitive editing ability}
    \vspace{-1.2em}
    \caption{The performance of MIND and SERAC~\cite{mitchell2022memory} on traditional cognitive editing benchmark and our proposed CogEdit.}
    \label{fig:intro_bar}
    \vspace{-15pt}
\end{figure}
Existing knowledge editing datasets primarily evaluate a model's cognitive capabilities but fail to effectively reflect its meta-cognitive abilities. To address this limitation, we introduce \textsc{\textbf{CogEdit}}. It is a novel benchmark designed to assess the meta-cognitive abilities of MLLMs during the knowledge editing process, which encompasses tasks of varying dimensions across diverse domains. In alignment with the three key levels outlined earlier, our benchmark evaluates models across three essential criteria: 

 % \begin{itemize}[leftmargin=*]
 % Counterfactual condition-driven Editing
    \textbf{Counterfactual-Driven Editing(Level 1)} – 
    When proposing a counterfactual assumption that alters the content of specific knowledge, the model should be \textbf{\underline{self-aware}}  that the change in the correctness of the knowledge is attributed to the introduction of the counterfactual condition.
    % The model should be \textbf{\underline{self-aware}} of when prior knowledge is incorrect, particularly in cases where the newly introduced knowledge is inherently counterfactual.
    % Overgeneralization Mitigation through Boundary Constraints
    
    \textbf{Boundary-Constraint Editing(Level 2)} – 
    The model should \textbf{\underline{monitor the utilization boundary}} of the new knowledge to prevent it from being overgeneralized to seemingly similar yet unrelated scenarios and mistakenly suppressing knowledge that should be activated.
    % The model should monitor contextual variations and apply \textbf{\underline{contextual regulation}} to identify scenarios where newly introduced knowledge is overly generalized, applying appropriate constraints to maintain specificity.
    % Robustness to noise during the editing process
    
    \textbf{Noise-Robust Editing(Level 3)} – 
    When confronted with noisy knowledge of uncertain accuracy, the model must critically assess its validity through \textbf{\underline{reflective thinking}}, determining whether it should replace existing ones.
With our newly designed benchmark, we perform a comprehensive evaluation of SOTA multimodal editing methods for meta-cognitive editing. As shown in Figure~\ref{fig:intro_bar}, though existing methods~\cite{mitchell2022memory} achieve strong performance on cognitive knowledge editing~\cite{cheng2023can}, they lack meta-cognitive capabilities including self-awareness, boundary monitoring, and reflective thinking.

% The human brain exhibits meta-cognitive abilities from a very young age, enabling efficient learning, creativity, and critical thinking in everyday life.  
To address these limitations and advance meta-cognitive knowledge editing, we introduce \textbf{MIND} (\textbf{M}eta-cognitive \textbf{IN}tegrated \textbf{D}ynamic Knowledge Editing), a framework that emulates the human knowledge editing process by mimicking how humans learn new information within their minds.
% ========================================== before
% Specifically, it first constructs a memory of meta-knowledge, composed of meta-declarative and meta-conditional knowledge, to facilitate self-awareness of knowledge updating.
% Secondly, it introduces game theory to monitor the activation boundary of new knowledge via the interaction between old and new knowledge. 
% Additionally, MIND incorporates a Label Refiner module to filter out noisy new knowledge and ensure the robustness of the editing process.
% As a result, MIND enables MLLMs to realize ``thinking about thinking'' by exerting self-control over its own cognitive processes.
% As shown in Figure~\ref{fig:intro_bar}, It attains strong performance on both traditional cognitive knowledge editing benchmark and our newly proposed meta-cognitive knowledge editing benchmark.

% ========================================== after
MIND begins by building a meta-knowledge memory, including both meta-declarative and meta-conditional components, to enable self-aware knowledge updates. It then applies game theory to monitor knowledge activation through old-new knowledge interactions and employs a label refiner to filter noisy inputs, enhancing robustness. This enables MLLMs to perform ``thinking about thinking'' by regulating their own cognitive processes. As shown in ~\autoref{fig:intro_bar}, MIND achieves strong performance on both standard and newly proposed meta-cognitive editing benchmarks.

Overall, our main contributions are three-fold:
% 其次，引入博弈理论通过新老知识的协同博弈来控制新知识的激活边界
% Additionally, MIND incorporates a Label Enhancement module to filter out 不正确的错误新知识干扰，确保了编辑过程的鲁棒性。
% MIND does not require altering the original model's weights; instead, it performs knowledge editing by selectively activating and inhibiting knowledge neurons. MIND constructs a mapping relationship using editing examples, enabling the suppression of erroneous knowledge and the activation of correct knowledge without affecting irrelevant weights. Additionally, MIND incorporates a Label Enhancement module that simulates the human brain’s ability to filter out irrelevant information, thereby enabling more precise identification of neurons corresponding to erroneous and correct knowledge.
\begin{itemize}
    \item We introduce CogEdit, a benchmark to evaluate meta-cognitive knowledge editing capabilities for MLLMs.
    \item We propose MIND, a framework that simulates human-like meta-cognitive learning with self-awareness, boundary monitoring, and reflective thinking.
    \item MIND achieves strong performance on both traditional cognitive knowledge editing benchmark and our newly developed meta-cognitive benchmark.
\end{itemize}

\section{Related Work}
% 1. multimodal editing的定义，一句话
% 2. benchmark层面，有哪些benchmark. 有哪几个指标. 客观评价衡量cognition,一句话概括什么是cognition,同时多样性有限。考虑到editing是meta-cognition的，我们提出了XXXwith多样性，来衡量123的能力。
% 3. 方法层面，现有方法包含intrinsic定义，external定义. 迎合cognition，没有具备meta-cognitive。对比，我们提出的XXX，人脑...
% 

% Multimodal Knowledge Editing is the process of modifying, updating, or correcting specific knowledge within a MLLM while preserving its overall performance across different modalities.
% 现有的multimodal editing benchmark衡量编辑后的效果 with Reliability, Generality and Locality metrics. 这些指标都衡量了编辑方法的cognition能力，即能够识别到旧知识被新知识来替换，但是多样性有限。Considering knowledge editing is meta-cognitive, we propose CogEdit with 多样性, 来衡量编辑方法是否有self-awareness, boundary monitoring and reflective thinking.
% 方法层面，现有的知识编辑方法包括直接对模型中存储的参数化只是进行改变的intrinsic knowledge editing 和保护模型原有参数，通过对编辑样本引入外来知识库进行编辑external knowledge resorting方法. 但是这两类方法都没有具备Meta-Cognitive的ability。因此，我们提出了MIND， a knowledge editing framework that simulates human-like meta-cognitive learning.

Multimodal Knowledge Editing is the process of modifying, updating, or correcting specific knowledge within MLLMs~\cite{liu2023llava, zhu2023minigpt, pan2024auto, li2023blip2bootstrappinglanguageimagepretraining, dai2023instructblipgeneralpurposevisionlanguagemodels, li2023fine, pan2025focusdiff} while preserving its overall performance across different modalities.
\begin{figure*}[t]
    \centering
    \includegraphics[width=0.97\textwidth]{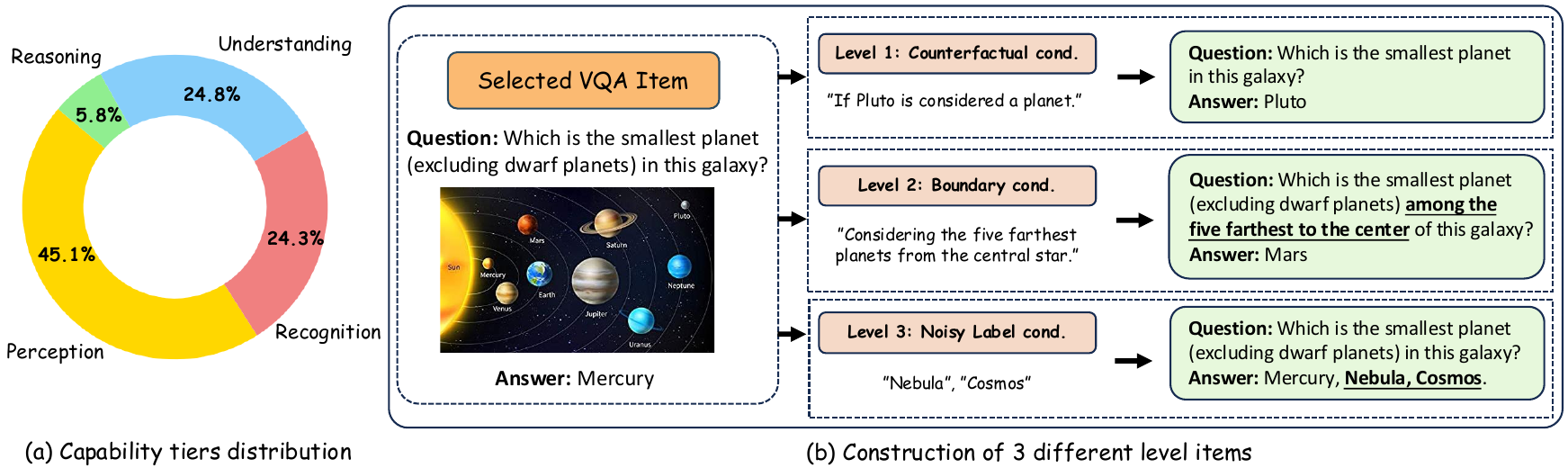}
    \vspace{-7px}
    \caption{(a) Distribution of problem difficulties in \textbf{CogEdit}. (b) The transformation process from an original VQA instance to edited instances at different levels.}
    \label{fig:dataset}
    \vspace{-10px}
\end{figure*}
Existing multimodal editing benchmarks~\cite{DBLP:conf/emnlp/0008TL0WC023,li2024mike, huang2024vlkeb, zhang2024mcmkefinegrainedmultimodalknowledge} evaluate the effectiveness of editing using the Reliability, Generality, and Locality metrics~\cite{huang2023transformer}. These metrics assess the cognitive capability of editing methods, specifically their ability to recognize when old knowledge is replaced by new one. However, they exhibit limited diversity. Considering that knowledge editing is inherently meta-cognitive, we propose \textbf{CogEdit}, which introduces diversity as an additional metric to evaluate whether an editing method possesses self-awareness, boundary monitoring, and reflective thinking.

From a methodological perspective~\cite{pan2025towards}, existing knowledge editing approaches can be categorized into two types: (1) Intrinsic Knowledge Editing~\cite{de2021editing, mitchell2021fast, meng2022locating, li2024pmet}, which directly modifies the model’s internal parameterized knowledge, and (2) External Knowledge Resorting~\cite{zheng2023can, mitchell2022memory, zhong2023mquake, wang2024wiserethinkingknowledgememory, chen2024lifelong}, which preserves the original model parameters while incorporating external knowledge sources through edited samples. However, neither of these approaches exhibits meta-cognitive capabilities. In contrast, we introduce \textbf{MIND}, a knowledge editing framework that simulates a human mind's meta-cognitive learning process.

\section{CogEdit Benchmark}

In this section, we propose \textbf{CogEdit}, a meta-cognitive multimodal  editing benchmark that encompasses \textbf{multiple capability levels} across \textbf{diverse domains}. 
Corresponding to the three levels of meta-cognition (\textit{i.e.}, self-awareness, boundary monitoring, reflective thinking), CogEdit evaluates multimodal editing across three essential criteria.
We give more details of CogEdit in Appendix A.

\paragraph{Task Definition.}
Assuming we have an MLLM parameterized by $\theta$ that map the
input $(i_e, x_e)$  to the prediction $y_o$, where $i_e$ refers
to the editing image input, $x_e$ refers to the editing
text prompt input and $y_o$ denote as the origin output.
Multimodal editing aims to change prediction from $y_o$ to $y_e$ after an updated $\theta_e$. On this basis, we introduce our triple-level editing subtasks, as illustrated in ~\autoref{fig:dataset}. A formal definition of the task and corresponding evaluation metrics can be found in~\autoref{appendix:dataset_details}.
 
\paragraph{Level 1: Counterfactual-Driven Editing.}
The first level of CogEdit focuses on evaluating the model's ability to demonstrate \textbf{self-awareness}. 
When a counterfactual assumption alters the correctness of specific knowledge, the model should recognize that the change in the correctness of that knowledge is due to the introduction of the counterfactual condition.
When these conditions are subsequently removed, the prior knowledge should be reactivated.
% Specifically, it assesses whether the model, after being edited with counterfactual conditions and a corresponding response, can accurately reactivate prior knowledge when those counterfactual conditions are removed.

% Editing with counterfactual knowledge involves leveraging scenarios under counterfactual conditions and evaluating the model’s performance in both counterfactual and original contexts.

% Formally, for a given instance $(i_e, x_e)$ (\textit{e.g.}, $i_e$ is an image of ``\textit{Solar System}''; $x_e$=``\textit{Which is the smallest planet in this galaxy?}'') with original correct answer $y$=``\textit{Mercury}'', 
% we introduce a counterfactual assumption $cf_e$=``\textit{Assume that Pluto is a planet, rather than a dwarf planet}'', which adjusts the correct answer to $y_e$=``\textit{Pluto}'' with the editing objective as $\mathbf{1}_{f(cf_e, i_e, x_e; \theta_e(cf_e, i_e, x_e)) = y_e}$ (we use \textbf{Fidelity} as the editing success rate).
% After editing, we remove the counterfactual assumption with $y_o$ restored to the correct answer, assessing whether the edited model can recall its prior correct knowledge when the assumption no longer holds: $\mathbf{1}_{f(i_e, x_e; \theta_e(cf_e, i_e, x_e))=y}$ (where \textbf{Adaptability} quantifies the success rate of answering correctly after removing the counterfactual assumption).

The task assesses a model's capability to incorporate counterfactual knowledge by measuring \textbf{Fidelity}—the accuracy under a counterfactual assumption—and \textbf{Adaptability}—the ability to recover original knowledge once the assumption is removed.

% ==================================================
% \subsubsection{Boundary-restricted editing}
% Since 在特定的场景和限制条件下，某些原本成立的知识可能会不成立
% 更精确的定位所需要用到的知识 needs the ability of understanding the reason why the new knowledge is correct, especially the scenarios and conditions where the new knowledge applies, rather than simply activating it in seemingly similar contexts.

\paragraph{Level 2: Boundary Constraint Editing.}
% The second level for CogEdit evaluation 测试了模型是否能够\textbf{monitor} the 时刻 and 场景 to apply 新知识，我们设计了一个和需要被edit sample相似但是添加了boundary-restricted场景，使得在这个场景下原有的答案失效。我们将这个指标记为Boundary Compliance，而把原有based场景下的编辑成功率记为Reliability.
% Formally, for a given instance \( p_d = (i_d, x_d) \), 我们通过在场景中添加一些限制\delta_{b} 来构造一个新的问题 \( p_b = \mathcal{C}(p_d) = (i_d, x_d) + \delta_{b}\)，并构造对应的正确答案 \( t_b \)。此时，先前编辑的知识应被抑制。 我们分别用Reliability和Compliance来衡量模型编辑的成功率和能否正确应用新知识。

% The second level of \textbf{CogEdit} evaluation assesses whether the model can \underline{\textbf{monitor}} the context and conditions under which new knowledge should be applied. To achieve this, we design scenarios that are similar to those requiring editing but introduce boundary restrictions, rendering the original answer invalid. We define this evaluation metric as \textit{Boundary Compliance}, while the success rate of editing under the original conditions is referred to as \textit{Reliability}. 

The second level of CogEdit evaluation assesses the model's ability of \textbf{boundary monitoring}, preventing the new
knowledge being overgeneralized to seemingly similar yet unrelated scenarios, and avoiding the erroneous suppression of knowledge that should be activated.

% We refer to this evaluation metric as \textit{Boundary Compliance}, while the success rate of editing under the original conditions is termed \textit{Reliability}.

% Formally, for the previously given instance \( (i_e, x_e) \) with editing target $y_e$=``\textit{Mercury}'', we conduct the conventional editing process to let the model remember the correct answer in the original context, with the editing objective as $\mathbf{1}_{f(i_e, x_e; \theta_e(i_e, x_e)) = y_e}$ (we use \textbf{Reliability} as the editing success rate).
% After editing, we construct a new problem instance by introducing boundary constraints  $b_e$=``\textit{Considering the five farthest planets from the central star}'', which limits the available candidate planets and adjusts the correct answer to $y'_e$=``\textit{Mars}'', in order to assess whether the edited model knows the boundary to utilize the edited knowledge rather than  indiscriminately activating it: $\mathbf{1}_{f(b_e, i_e, x_e; \theta_e(i_e, x_e))=y'_e}$ (where \textbf{Compliance} quantifies the success rate of answering correctly after adding the boundary constraint assumption).

This task evaluates a model’s ability to apply edited knowledge appropriately by measuring \textbf{Reliability}—its accuracy in the original context after editing—and \textbf{Compliance}—its ability to adapt its responses under boundary constraints without overgeneralizing the edited knowledge.

\paragraph{Level 3: Noise-Robust Editing.}

% Noisy label editing introduces noise into the labels that require knowledge editing, aiming to evaluate the model's ability to handle ambiguity during the editing process. Specifically, for a label that needs to be edited, noise is added to the label text, simulating scenarios where the target knowledge is unclear or partially corrupted. The objective is to assess the model's capability to discern and correctly apply the edit under such noisy conditions.
% The third level in CogEdit is testing whether model can \underline{\textbf{think reflectively}} and critically about the knowledge it is editing. This is done by introducing noise into the labels that require knowledge editing, simulating scenarios where the target knowledge is unclear or partially corrupted. The objective is to assess the model's capability to discern and correctly apply the edit under such noisy conditions. For a given instance \( p_d = (i_d, x_d) \), we construct a noisy editing instance by adding noise to the target knowledge \( t_d \), forming \( p_n = (i_d, x_d) \), \( t_n = t_d + t_{noise}(k) \), where \( k \) represents the number of noisy elements added to the label.
% Thus, we introduce a new evaluation metric, \textit{Clarity@K}, to quantify the model's robustness and adaptability to noisy inputs.

The third level in CogEdit evaluates whether the model can engage in \textbf{reflective thinking}. When confronted with noisy or uncertain knowledge, the model should critically assess its validity and determine whether it should be accepted as a replacement for existing knowledge.

This task measures a model’s ability to extract and retain useful information from noisy updates. Specifically, \textbf{Clarity@K} quantifies the model’s robustness by evaluating its accuracy as the level of injected noise increases, with \( K \) denoting the number of distracting elements.

\paragraph{Data Collection.}
% 这里用一张图来展示不同的capability level和domain
To ensure diversity in data sources, we curated a total of 1,174 VQA questions from various multimodal QA datasets~\cite{hudson2019gqanewdatasetrealworld, marino2019okvqavisualquestionanswering}. These questions span various specialized domains, including sports, arts, science, everyday activities and so on. The selection process also considered different capability tiers, categorized from simplest to most complex: \textbf{Perception}, \textbf{Recognition}, \textbf{Understanding}, and \textbf{Reasoning}. We use open-source models~\cite{chen2024internvlscalingvisionfoundation, grattafiori2024llama3herdmodels, li2024llava, wang2024qwen2} to help us generate and refine the data instances.
% \begin{itemize}
%     \item \textbf{Perception:} Scene understanding, instance identity, attribute recognition, and spatial localization.  
%     \item \textbf{Multimodal Recognition:} Text recognition.
%     \item \textbf{Multimodal Understanding:} Visual reasoning and visual referring expressions.  
%     \item \textbf{Human-Centric Cascading Reasoning:} Logical reasoning, numerical reasoning, and visual mathematics.  
% \end{itemize}
% To ensure the reliability and accuracy of the curated data, we conducted a two-pass quality control process. For the first pass, we utilized open-source models\cite{chen2024internvlscalingvisionfoundation, grattafiori2024llama3herdmodels} to automatically validate and refine the data. In the second pass, a manual review was performed to further ensure correctness in order to remove ambiguous or poorly phrased entries. This combination of automated and manual methods ensured consistency with the intended domain and capability levels.

% \paragraph{Quality Control}
% To ensure the reliability and accuracy of the curated data, we conducted a two-pass quality control process. For the first pass, we utilized open-source models\cite{chen2024internvlscalingvisionfoundation, grattafiori2024llama3herdmodels} to automatically validate and refine the data. In the second pass, a manual review was performed to further ensure correctness in order to remove ambiguous or poorly phrased entries. This combination of automated and manual methods ensured consistency with the intended domain and capability levels.

\section{Approach: Meta-Cognitive Editing}

% We propose a meta-cognitive multimodal editing method, \textbf{MIND} (\textbf{M}eta-cognitive \textbf{IN}tegrated \textbf{D}ynamic knowledge editing), to overcome the limitations of cognitive-based editing in precisely locating and modifying knowledge. \textbf{MIND} 通过在编辑过程中保留原有知识的激活通路，根据场景特征指导编辑和更新，并提取出编辑样例中的有效信息，which integrates a self-aware meta-cognitive memory\ref{subsec:method_aware}, a game-theoretic regulation mechanism\ref{subsec:method_monitor}, and a Meta-Label Refiner\ref{subsec:method_reflective}, enabling meta-cognitive knowledge editing in a manner similar to how the human mind learns and adapts.

\begin{figure*}[tbp]
    \centering
    \includegraphics[width=0.94\textwidth]{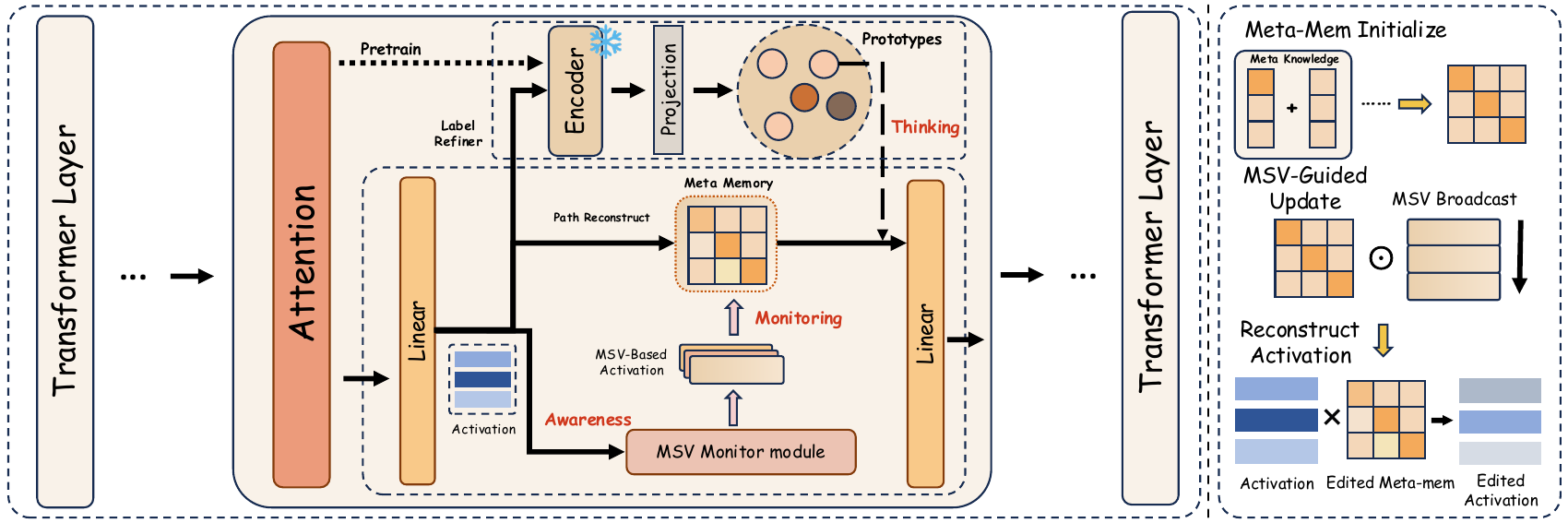}
    \vspace{-11px}
    \caption{The MIND framework unifies meta-memory, monitoring mechanisms, and label refinement for meta-cognitive multimodal knowledge editing.}
    \label{fig:method}
    \vspace{-16px}
\end{figure*}

We propose a meta-cognitive multimodal editing method, \textbf{MIND} (\textbf{M}eta-cognitive \textbf{IN}tegrated \textbf{D}ynamic knowledge editing), to overcome the limitations of cognitive-based editing, as shown in ~\autoref{fig:method}. \textbf{MIND} retains the activation pathways of the original knowledge during the editing process, guiding the editing and updating based on scene features, and extracting relevant information from the editing examples. It integrates a self-aware meta-cognitive memory(\ref{subsec:method_aware}), a game-theoretic monitoring mechanism(\ref{subsec:method_monitor}), and a meta-label refiner(\ref{subsec:method_reflective}), enabling meta-cognitive knowledge editing in a manner similar to how the human mind learns and adapts.

\subsection{Self-Aware Meta Knowledge}
\label{subsec:method_aware}
% 或许可以提到Lifelong，也可以直接在exp里说
% Knowledge分为几种类型的，其中Declarative Knowledge在模型的Pretrain和微调过程中，会被学习在不同参数的分布中
% 引用https://ctl.utexas.edu/metacognition

% - Declarative Meta-Knowledge（陈述性知识）：对记忆内容的了解，例如某些信息更容易记住。
% - Procedural Meta-Knowledge（程序性知识）：对如何更好地记忆的策略了解，例如使用记忆术或复习间隔策略。
% - Conditional Meta-Knowledge（条件性知识）：知道何时、何地、为何使用某种记忆策略。

% \paragraph{Editing Feed-Forward Layers}  

% The keys correspond to textual patterns in the training data, while the values determine the output distribution over the vocabulary. The inner hidden states can be interpreted as query vectors that regulate neuron activation and inhibition. 

% Thus, a feed-forward layer, which applies two linear transformations with an interleaved non-linear activation, can be expressed as:  

% \begin{equation}  
%     \text{FFN}(x) = Linear_V(Linear_K(x))
% \end{equation}  

% where \( Linear_V \) and \( Linear_K \) represent linear transformations.
% % and \( \text{ACT} \) denotes a non-linear activation function.  

% Thus, after passing through the \( Linear_K \) layer, the data is transformed into an activation mapping for \( Linear_V \). In the process of knowledge editing, it is not necessary to modify the model’s existing parameterized knowledge; instead, knowledge editing can be achieved by selectively modifying the activation and inhibition of specific neurons.

\paragraph{Meta-Memory Construction.}  
Feed-forward layers in transformer models, which constitute a significant portion of the model's parameters, function as key-value memory structures \cite{geva2020transformer}. Thus make it possible to edit knowledge through editing feed-forward layers.

Since activation and inhibition mappings influence the model's activation and forwarding logic, we define a Meta-Memory~\cite{nelson1990metamemory, cavanaugh1982metamemory} projection matrix \( \texttt{Mem} \in \mathbb{R}^{d' \times d'} \) which consists of the meta-knowledge units \( m_i \), which is a mapping function that extracts new activation values from original activation values. For a feed-forward layer with an inner hidden dimension of \( d' \), the meta-memory \texttt{Mem} can be represented with: 
% \begin{equation}  
%     Q_o = P(Q_i) = [f_1(Q_i), f_2(Q_i), \dots, f_d(Q_i)]
%     % Input Q_i: A vector of dimension d
%     % Output Q_o: A vector of dimension d
%     % Transformation: Each dimension of the output vector extracts features from the input using a function \( f_i \).
% \end{equation} 
% \begin{equation}  
%     q_o = P(q_i) = \operatorname{stack}_{i=1}^{d} \bigl( f_i(q_i) \bigr)
% \end{equation} 
% Input q_i: A vector of dimension d
% Output q_o: A vector of dimension d
% Transformation: Each dimension of the output vector extracts 
% \vspace{-13px}
\begin{equation} 
    q_{output} = \texttt{Mem}(q_{input}) = \bigl[\, m_i(q_{input_i}) \,\bigr]_{i=0}^{d'-1}
\end{equation} 
\begin{equation}  
    m_i = \operatorname{split}(\texttt{Mem})_i = \texttt{Mem}[:, i] \quad \forall i \in \{0, 1, \dots, d'-1\}
\end{equation} 
where \( q_i \) and \( q_o \) are original and new activation values of dimension \( d' \), respectively.

Meta-knowledge is parameterized and stored in meta-memory, and it can be categorized into different levels, including \textbf{meta-declarative knowledge}  \( m_i^{\text{decl}} \) and \textbf{meta-conditional knowledge} \( m_i^{\text{cond}} \). 

% \textbf{Meta-declarative knowledge} refers to a language model's awareness and understanding of its own knowledge and structure. While normal declarative knowledge in large language models encompasses facts and data about the world, linguistic rules, and domain-specific information—learned during pretraining from large-scale textual sources such as books and websites—meta-declarative knowledge enables a model to recognize what it knows, its knowledge limitations, and how its responses are generated or retrieved. For example, a language model may learn that ``human beings need oxygen to survive'' or ``the sky is blue'' during pretraining, but meta-declarative knowledge allows it to assess the reliability and origin of such statements.

% \textbf{Meta-declarative knowledge} refers to a language model's self-awareness of what it knows, how that knowledge is structured, and its limitations. Unlike regular declarative knowledge—which includes factual and linguistic information learned during pretraining—meta-declarative knowledge enables the model to evaluate the reliability and origin of its responses.

\textbf{Meta-declarative knowledge} refers to a language model's self-awareness of its knowledge, structure, and limitations. It goes beyond storing facts by allowing the model to assess how reliable its responses are and where they come from.

% \textbf{Meta-conditional knowledge}, on the other hand, pertains to the awareness of the conditions under which specific knowledge or skills should be applied. It identifies the contextual factors that determine the relevance and appropriateness of utilizing certain knowledge or strategies. This form of meta-knowledge facilitates adaptability and informed decision-making by allowing a model to evaluate when and how to apply particular information effectively in dynamic environments.

\textbf{Meta-conditional knowledge} refers to a model's awareness of when and under what conditions specific knowledge should be applied. It helps the model adapt by evaluating the relevance and appropriateness of its knowledge in varying contexts. This meta-knowledge helps the model adapt and make informed decisions by guiding when and how to apply information effectively.

% 初始时，\( m_i^{\text{decl}} = \mathbb{1}_{j = i} \) and \( m_i^{\text{cond}} = all \; \mathbf{0}\)，表示初始情况下，meta-memory会维持原有事实性的declarative knowledge，但还未学习到conditional knowledge。

% Initially, \( m_i^{\text{decl}} \) and \( m_i^{\text{cond}} \) are set with the following values, indicating that the meta-memory retains the original declarative knowledge but has not yet learned conditional knowledge.
% \begin{equation}
%     \begin{aligned}
%          m_i^{\text{decl}} &= 1 \quad \text{if} \; i = \text{th index}, \; \text{else} \; 0 \\
%         m_i^{\text{cond}} &= \mathbf{0} \quad \text{for all} \; i
%     \end{aligned}
% \end{equation}

Initially, \( m_i^{\text{decl}} = 1 \) if \( i \) equals the target index, and \( m_i^{\text{decl}} = 0 \) otherwise; meanwhile, \( m_i^{\text{cond}} = \mathbf{0} \) for all \( i \), indicating that the meta-memory retains the original declarative knowledge while no conditional knowledge has been learned yet.

% =============================

% \paragraph{Atomic Operations of Meta Knowledge}
% % 编码、储存、检索、遗忘、重建与重组、认知调节、知识整合和适应性调整
% Meta-knowledge operations can primarily be categorized into the following types, based on the operations related to knowledge in psychology.
% \begin{itemize}
%     \item \textbf{Encoding}: The process of transforming external information into mental representations that can be stored in memory. It involves attention and meaningful processing.
%     \item \textbf{Storage}: The act of maintaining encoded information in memory over time. This includes both short-term and long-term memory storage mechanisms.
%     \item \textbf{Retrieval}: The process of accessing and bringing stored information into consciousness for use in cognitive tasks or decision-making.
%     \item \textbf{Forgetting}: The inability to recall or access stored information. This can occur due to memory decay, interference, or lack of retrieval cues.
%     % \item \textbf{Knowledge Integration}: The combination of new information with existing knowledge networks, forming a more complex and interconnected structure of understanding.
% \end{itemize}

% ================================================
\subsection{Game Theory based Meta-Memory Monitoring}
\label{subsec:method_monitor}

% 1.Most knowledge editing methods focus on editing the feed-forward layers of the model.
% TODO: from https://arxiv.org/pdf/2104.08696 perspective of view
% --------------------
% \paragraph{Preliminary.}
% \textbf{Shapley value} is a solution concept in cooperative game theory. It represents the \textbf{marginal contribution} of each player to the overall outcome in a cooperative game. For a cooperative game defined by a set of players \( N \) and a characteristic function \( v: 2^N \to \mathbb{R} \), the Shapley value for a player \( i \in N \) is defined as:

% % \vspace{-20pt}
% \begin{equation}
% \phi_i(v) = \sum_{S \subseteq N \setminus \{i\}} \omega(S, N) \cdot \left[ v(S \cup \{i\}) - v(S) \right]
% \end{equation}

% The term \( \omega(S, N) = \frac{|S|! \, (|N| - |S| - 1)!}{|N|!} \) represents the weight assigned to the marginal contribution of player \( i \) to the subset \( S \), where \( S \) is a subset of players excluding player \( i \). Here, \( |S| \) denotes the size of the subset \( S \), \( |N| \) is the total number of players, and \( v(S) \) refers to the value of the coalition \( S \), as defined by the characteristic function \( v \).

% \paragraph{Neuron Shapley Value}
% 2. The importance of each neuron in the hidden layer can be learned through the competition mechanism.

Since most knowledge is stored in feed-forward layers, and meta-memory manipulation provides a mechanism to control the knowledge that needs to be activated, it is crucial to understand the importance of each meta-memory unit \( m_i \) in the hidden layer within a specific knowledge context. To address this, we introduce the concept of the \textbf{Meta-memory Shapley Value (MSV)}~\cite{roth1988introduction, winter2002shapley}, which quantifies the marginal contribution of each unit to the overall editing performance.In alignment with the intermediate activation values \( q \) between two layers of the FFN, MSV can be defined as follows:
\begin{equation}
    \phi{(Mem, q)} = \sum_{s \subseteq \texttt{Mem} \setminus \{i\}} \left( v(s \cup \{i\}, q) - v(s, q) \right)
\end{equation}
where Mem denotes the set of all meta-memory units, and \( s \) represents a subset of units excluding \( m_i \). The term \( v(s, q) \) quantifies the performance or contribution of the subset \( s \) of meta-memory units for the query \( q \), while \( v(s \cup \{i\}, q) \) denotes the performance or contribution of the subset \( s \) combined with the unit \( m_i \) for the same query.

% This formulation measures the total marginal contribution of \( N_i \) across all possible subsets \( s \) of neurons in the context of query \( q \).

% By aggregating these contributions, we can determine the importance of each neuron for the query \( q \) and adjust the projection matrix \( P \) accordingly to fine-tune the model's behavior.

% TODO: explain3 use N to represent Neuron and d to represent the dimension of the hidden layer, the projection matrix P can be represented as a d x d matrix, where each row of P represents the importance of each neuron in the hidden layer. Different weights of cooperation can cause different editing effects, and the importance of each neuron can be learned through the competition mechanism.
% Write a formula to represent the Neuron Shapley value in the feed-forward layer(using sigma to represent marginal benefit). Explain the importance of each neuron in the hidden layer can be learned through the competition mechanism.

Due to the computational complexity of calculating the marginal contribution of each memory unit, we employ an \textbf{MSV Monitoring module} with MLP as the approximation~\cite{lin2024teamlora} of MSV. The MLP is trained to predict the \textbf{MSV} according to the current input query \( q \). The marginal contribution of the \( i \)-th unit of meta-memory is computed as follows:
% Apply softmax normalization to the Neuron Shapley value
\begin{equation}
    % \text{MSV}(q)_i <- \text{Softmax}(\phi(N, q))_i
    \phi_i(\texttt{Mem}, q) \leftarrow \text{Softmax}(\text{Shapley}(m_i, q))
\end{equation}

% =========================================

% \paragraph{Guided Meta-Memory Regulation}  
% Leveraging the extracted information from Neuron Shapley Value (MSV) facilitates a more effective regulation of the knowledge update process. Since MSV represents the significance of each neuron’s corresponding knowledge within the given context, its weights can be utilized as a guiding mechanism to update the \textbf{Knowledge Path P}.

% Assume that hidden state is d, intermediate hidden size in FFN is d'.
% The meta-memory module mainly contains a projection matrix M ∈ d' x d', a neuron shapley competing module. 
% For an arbitrary input x, first the Key matrix is calculated as Linear_K(x),即original knowledge path. During editing with MIND, the projection matrix M is inserted after original knowledge path, 即unbalanced knowledge path q_u = Linear_K(x) * M
% Meanwhile, the neuron shapley competing module is used to extract the contextual information from the input x, 即MSV(x) = \text{Softmax}(\phi{(N, x)})

% 为了更清楚地明白什么时候新编辑知识需要被应用，MSV is used to guide the meta-memory update process. Since each column in knowledge path represents the activation values, knowledge path is splitted into d' columns first.

% q_u_i = split(q_u)_i = [q_u的第i列]
% Then the update is guided with the MSV,
% where balanced knowledge path q = [q_u_0 * MSV_0, q_u_1 * MSV_1 ... q_u_d' * MSV_d']

% \paragraph{Guided Meta-Memory Regulation}  
Leveraging the extracted information from MSV facilitates a more effective monitoring of the knowledge update process. 

To effectively determine when newly edited knowledge should be applied, MSV is utilized to guide the meta-memory update process, enabling a structured monitoring of meta-memory, which operates as follows:
\begin{equation}
    q_r = \bigl[\, (q_i \cdot m_i) \odot \phi_i(\texttt{Mem}, q) \,\bigr]_{i=0}^{d'-1}
\end{equation}
% Then the reconstructed meta-memory based activation is used to query the neurons in \(Linear_V\), forming the layer output as \( Linear_V(q) \) and the regulated feed-forward layer as:

% FIXME: The following paragraph will set to appendix
% \begin{equation}
%     \text{FFN}_{edit}(x) = \text{Linear}_V (\bigl[\, (q \cdot M)_i \cdot \text{Softmax}(\phi(N, q))_i \,\bigr]_{i=0}^{d'-1})  where \; q=\text{Linear}_K(x)
% \end{equation}  
% Thus forming the regulated feed-forward layer as:

% \begin{equation}
%     \begin{aligned}
%         \text{FFN}_{r}(x) &= \text{Linear}_V \left( \left[ (q_i \cdot m_i) \circ \phi_i(\texttt{Mem}, q) \right]_{i=0}^{d'-1} \right) \\
%         \text{where} \; q &= \text{Linear}_K(x)
%     \end{aligned}
% \end{equation}

% ================================================
\renewcommand{\arraystretch}{1}
\newcommand\resulttablefontsize{\fontsize{7.7pt}{9.24pt}\selectfont}
\newcommand{\first}{\colorbox{blue!25}}
\newcommand{\second}{\colorbox{blue!10}}

\begin{table*}[t]
	\centering
    \caption{Main results on \textbf{CogEdit}, including Counterfactual-Driven Editing, Boundary Constraint Editing, and Noise-Robust Editing.}
    \vspace{-8px}
	\resulttablefontsize
	\setlength{\tabcolsep}{4pt}
	\resizebox{2\columnwidth}{!}{
	\begin{tabular}{l c c c | c c | c c }
		\toprule
		\multicolumn{2}{c}{} & \multicolumn{2}{c}{\textsc{Counterfactual}} & \multicolumn{2}{c}{\textsc{Boundary Editing}} & \multicolumn{2}{c}{\textsc{Noisy Editing}} \\
		\cmidrule(r){3-4} \cmidrule(r){5-6} \cmidrule(r){7-8}
        & \multicolumn{1}{c}{Method} 
        & \multicolumn{1}{c}{Fidelity$\uparrow$} 
        & \multicolumn{1}{c}{Adaptability$\uparrow$} 
        & \multicolumn{1}{c}{Reliability$\uparrow$} 
        & \multicolumn{1}{c}{Compliance$\uparrow$} 
        & \multicolumn{1}{c}{Clarity@2$\uparrow$} 
        & \multicolumn{1}{c}{Clarity@4$\uparrow$}  \\ 
		\hline
		\multicolumn{1}{l}{\scriptsize \textcolor{darkgray}{}} & \multicolumn{6}{c}{\textbf{\small MiniGPT-4}} & \multicolumn{1}{r}{\scriptsize \textcolor{darkgray}{Size: 7.3B}} \\
		\hline
		\multirow{2}{*}{Base Methods} 
            & FT (vision block) & 55.08 & 16.72 & 52.56 & 23.20 & 19.29 & 17.84 \\
            & FT (last layer) & 67.49 & 15.85 & 60.93 & 24.66 & 17.66 & 14.12 \\
		\hline
		\multirow{5}{*}{Cognitive Editing} 
            & T-Patcher & 73.16 & 19.43 & 62.15 & 24.72 & 27.76 & 25.44 \\
            & IKE & \textbf{99.66} & 13.67 & \textbf{99.92} & 15.72 & 14.98 & 10.72 \\
            & MEND & 60.55 & 20.88 & 73.92 & 33.10 & 10.43 & 3.50 \\
            & SERAC & 99.26 & 29.98 & 99.68 & 40.49 & 30.88 & 30.83 \\
            & WISE & 76.29 & 27.63 & 79.33 & 31.00 & 29.33 & 29.01\\
		\hline
        Meta-Cognitive Editing 
        & \textbf{MIND (Ours)} & 98.16 & \textbf{34.31} & 88.41 & \textbf{43.21} & \textbf{34.98} & \textbf{34.87} \\
		\hline
		\multicolumn{1}{l}{\scriptsize \textcolor{darkgray}{}} & \multicolumn{6}{c}{\textbf{\small LLaVA}} & \multicolumn{1}{r}{\scriptsize \textcolor{darkgray}{Size: 7.06B}} \\
		\hline
		\multirow{2}{*}{Base Methods} 
            & FT (vision block) & 30.97 & 18.60 & 24.27 & 26.47 & 19.50 & 18.95 \\
            & FT (last layer) & 71.66 & 28.87 & 45.69 & 21.98 & 21.98 & 18.20 \\
		\hline
		\multirow{5}{*}{Cognitive Editing} 
            & T-Patcher & 82.55 & 33.62 & 77.64 & 29.39 & 30.33 & 30.69 \\
            & IKE & 78.44 & 31.87 & 76.51 & 28.26 & 34.68 & 39.50 \\
            & MEND & 85.36 & 47.44 & 94.90 & 44.84 & 4.98 & 4.98 \\
            & SERAC & 98.84 & 52.03 & \textbf{99.72} & 53.20 & 53.42 & 47.88 \\
		    & WISE & 78.72 & 37.39 & 65.67 & 36.23 & 47.66 & 50.28 \\
		\hline
        Meta-Cognitive Editing 
        & \textbf{MIND (Ours)} & \textbf{99.87} & \textbf{56.47} & 99.33 & \textbf{59.08} & \textbf{60.86} & \textbf{58.92} \\
		\bottomrule
	\end{tabular}
	}
    \vspace{-1em}
    \label{tab:cog_main}
\end{table*}

% \subsection{Results on MMEdit Dataset}
\begin{table*}[t]
\centering
\caption{Main results on cognitive benchmark \textbf{MMEdit} }
\vspace{-8px}
\resulttablefontsize
%\scriptsize
%\footnotesize
\setlength{\tabcolsep}{4pt}
%\hfill{}
\resizebox{2\columnwidth}{!}{
\begin{tabular}{l c c c c c c | c c c c c }

%\toprule
\toprule
\multicolumn{2}{c}{} & \multicolumn{5}{c}{\textsc{Editing VQA}}  & \multicolumn{5}{c}{\textsc{Editing Image Caption}}  \\
    \cmidrule(r){3-7}  \cmidrule(r){8-12} 
	& \multicolumn{1}{c}{Method} &  \multicolumn{1}{c}{Reliability}$\uparrow$  & \multicolumn{1}{c}{T-Generality}$\uparrow$  & \multicolumn{1}{c}{M-Generality}$\uparrow$  & \multicolumn{1}{c}{T-Locality }$\uparrow$ & \multicolumn{1}{c}{M-Locality}$\uparrow$ & \multicolumn{1}{c}{Reliability}$\uparrow$ & \multicolumn{1}{c}{T-Generality}$\uparrow$ & \multicolumn{1}{c}{M-Generality}$\uparrow$ & \multicolumn{1}{c}{T-Locality}$\uparrow$ & \multicolumn{1}{c}{M-Locality}$\uparrow$ \\
    \hline
	\multicolumn{1}{l}{\scriptsize \textcolor{darkgray}{}} & \multicolumn{9}{c}{\textbf{\small MiniGPT-4}} & \multicolumn{1}{r}{\scriptsize \textcolor{darkgray}{Size: 7.3B}} \\
	\hline
        \multirow{2}{*}{Base Methods} 
        % & Base Model & 0.0 & 0.0 & 0.0 & 100.0 & 100.0 & 0.0 & 0.0 & 0.0 & 100.0 & 100.0  \\
        & FT (vision block) & 36.3 & 0.3 & 10.9 & 100.0 & 9.3 & 3.1 & 0.0 & 2.2 & 100.0 & 8.6  \\
        & FT (last layer) & 70.1 & 65.7 & 63.9 & 72.6 & 65.8 & 67.4 & 65.1 & 62.8 & 63.5 & 52.7 \\
        \hline
        \multirow{4}{*}{Cognitive Editing} 
        % & Knowledge Editor & 91.8 & 89.0 & 60.8 & 96.9 & 67.8 & 96.6 & 67.8 & 57.4 & 97.3 & 64.4 \\
        & T-Patcher & 83.0 & 68.2 & 66.0 & 84.8 & 82.0 & 83.8 & 72.3 & 67.7 & 93.9 & 83.6  \\
        & IKE & 100.0 & 94.9 & 90.5 & 50.3 & 3.7 & 90.9 & 81.6 & 88.5 & 52.2 & 4.7 \\
        & SERAC & 87.7 & 87.6 & 85.9 & 97.5 & 14.2 & 91.8 & 91.4 & 91.0 & 97.9 & 7.2  \\
        & MEND &  98.8 & 98.6 & 82.2 & 98.2 & 81.1 & 96.6 & 96.1 & 76.3 & 98.4 & 75.3 \\
        & WISE & 91.8 & 90.4 & 87.1 & 99.9 & 86.3 & 95.8 & 91.7 & 88.8 & 100.0 & 81.5 \\
	\hline
     Meta-Cognitive Editing   & \textbf{MIND(Ours)} & 97.0 & 95.2 & 91.7 & 93.6 & 72.2 & 93.2 & 90.7 & 81.3 & 99.6 & 84.3 \\

	\hline
	\multicolumn{1}{l}{\scriptsize \textcolor{darkgray}{}} &\multicolumn{9}{c}{\textbf{\small LLaVA}}  & \multicolumn{1}{r}{\scriptsize \textcolor{darkgray}{Size: 7.06B}} \\
	\hline
	\multirow{2}{*}{Base Methods} 
    % & Base Model & 0.00 & 0.00 & 0.00 & 100.0 & 100.0 & 0.00 & 0.00 & 0.00 & 100.0 & 100.0  \\
        & FT (vision block) & 63.4 & 58.6 & 53.7 & 92.5 & 66.7 & 48.3 & 49.4 & 45.1 & 99.8 & 93.6 \\\
        & FT (last layer) & 62.5 & 61.9 & 59.9 & 64.8 & 32.9 & 73.4 & 72.7 & 67.9 & 99.0 & 91.4 \\

        \hline
	\multirow{5}{*}{Cognitive Editing} 
        % & Knowledge Editor & 85.2 & 72.3 & 59.3 & 83.5 & 58.6 & 77.4 & 54.8 & 57.4 & 77.3 & 54.1  \\
        & T-Patcher & 87.8 & 81.5 & 78.3 & 67.5 & 47.6 & 85.3 & 87.2 & 81.0 & 75.4 & 44.2  \\
        & IKE & 95.4 & 86.2 & 79.1 & 52.6 & 17.3 & 92.5 & 84.3 & 76.6 & 52.5 & 15.6 \\
      & SERAC & 96.4 & 94.3 & 87.2 & 91.5 & 23.5 & 97.5 & 96.9 & 97.6 & 87.6 & 29.2 \\
      & MEND & 93.7 & 92.3 & 91.9 & 91.0 & 87.2 & 94.8 & 92.2 & 89.6 & 91.3 & 90.3 \\
    & WISE & 87.0 & 80.3 & 73.5 & 100.0 & 81.5 & 94.6 & 87.6 & 70.9 & 99.9 & 78.3 \\
        % \midrule
        \hline
   Meta-Cognitive Editing   & \textbf{MIND (Ours)} & 97.5 & 96.9 & 92.5 & 91.2 & 76.1 & 96.9 & 96.8 & 92.7 & 95.6 & 79.2 \\
      % & MIND (Ours) & 97.41 & 96.85 & 92.46 & 91.79 & 64.37 & 96.92 & 96.79 & 92.74 & 95.53 & 67.15 \\

	\bottomrule

\end{tabular}
}
%\hfill{}
\label{tab:mmedit}  
\vspace{-1em}
%\vspace{-17pt}
\end{table*}

% \input{sec/results/lifelong_results}
% ================================================

\subsection{Reflective based Label Refinement} \label{subsec:method_reflective}

\paragraph{Prototype-Based Meta-Label Refiner.} % Disambiguation  
Inspired by \cite{wang2022pico+}, we incorporate concepts from the partial label learning framework and introduce a reflective memory module that stores learned prototype knowledge associated with different label categories. This module supports reflective thinking by updating prototype representations based on new labels. Initialized with pre-trained weights, it adaptively refines specific prototypes during editing using edit labels. The encoder processes activation values $q$ to capture contextual understanding. Using these embeddings, contextual label prototypes are retrieved from the prototype bank \(\mathcal{P} = \big\{\mathbf{m}_k \big\}_{k=1}^K\), where \(\mathbf{m}_k\) represents embeddings encapsulating relevant label-related knowledge. The retrieved contextual label knowledge \(p\) is then projected with a learnable projection matrix \(\mathbf{W}_p\) to get meta-label info, then serves as a shifting vector and is combined with the meta-memory monitored activation values \(q_r\) to produce the refined representation: 
\begin{equation}
q_{\text{refined}} = (1-\beta) q_r + \beta \mathbf{W}_p p
\end{equation}
where \(\beta\) controls the contribution of the monitored activation and the meta-label-based shift. The incorporation of reflective thinking ensures that the model does not passively store information but actively refines and adapts its learned representations based on prior knowledge and newly observed samples.  

\paragraph{Supervised Contrastive Pre-Training.}  
To further improve robustness of reflective-based label refinement, we employ contrastive learning~\cite{NEURIPS2020_d89a66c7, tian2020makes} during refiner pre-training. This process explicitly integrates generated wrong labels as noisy labels. By allowing the model to reflect on its past errors, contrastive learning optimizes the parameters of the prototype bank module, reinforcing its ability to distinguish between useful and misleading knowledge.

\section{Experiment}
% In this section.We first验证了在our meta-cognitive editing benchmark上MIND的编辑能力，结果展示在~\ref{subsec:5.2},同时出于公平考虑，我们也验证了在traditional cognitive benchmark中的能力，结果在\ref{subsec:5.3}中。此外，在lifelong editing的场景下，我们也验证了MIND的性能，结合深入的分析证明了MIND拥有meta-cognitive ability。

In this section, we first evaluate the editing capability of \textbf{MIND} on our meta-cognitive editing benchmark, with results presented in~\ref{subsec:5.2}. We also assess its performance on a traditional cognitive benchmark, as shown in~\ref{subsec:5.3}. These experiments, along with an in-depth analysis~\ref{subsec:5.5}, collectively demonstrate its meta-cognitive ability. Furthermore, we evaluate \textbf{MIND} in a lifelong editing scenario in ~\autoref{appendix:lifelong}.

% 就我们首先来验证我们的模型在我们这个Co editmark上的结果，In section, 比如说5.1，然后我们再去证明它的常规的班，那上的这个有效性就是infe几点几。然后我们进一步的讨论什么什么，相当于就是把你的这一个整个section的一个逻辑说一下。
\subsection{Experiment Setup}

\paragraph{Datasets.}
To evaluate the meta-cognitive editing capabilities of the model, we conduct experiments on two benchmarks: CogEdit and MMEdit. These experiments aim to demonstrate that our method exhibits both meta-cognitive and standard cognitive abilities. See more details in ~\autoref{appendix:implementation_details}.
% \paragraph{Baselines}
% In this study, we selected several commonly used multimodal model editing methods as baselines for comparison. Specifically, we included traditional meta-learning approaches frequently applied in multi-modal knowledge editing, such as MEND\cite{mitchell2021fast}, SERAC\cite{mitchell2022memory}, and Finetune. These methods were chosen to provide a comprehensive benchmark against which the performance of our proposed method could be evaluated. By including both established techniques and novel approaches, we aim to demonstrate the effectiveness of our method in comparison to existing state-of-the-art solutions. 实验在MiniGPT4和LLaVa两种不同架构的MLLM上进行，分别代表了query-transformer based and direct projection based MLLMs.
\paragraph{Baselines.} 
In this study, we selected several cognitive-based widely adopted multimodal model editing methods as baselines, including Transformer-Patcher~\cite{huang2023transformer}, IKE~\cite{zheng2023can}, MEND~\cite{mitchell2021fast}, SERAC~\cite{mitchell2022memory}, WISE~\cite{wang2024wiserethinkingknowledgememory}, and standard fine-tuning. Experiments were conducted on two distinct MLLM architectures: MiniGPT-4~\cite{zhu2023minigpt}, representing Qformer-based models~\cite{li2023blip}, and LLaVA-v1.5~\cite{liu2023llava}, representing projection-based models.

% 三种编辑的方法：Additional params、locate then edit、meta learning

\subsection{Results on CogEdit Benchmark}
\label{subsec:5.2}
% 结果在上一页的表格中

% FT(backbone)
% FT(vision)
% MEND
% SERAC
% IKE
% MetaKE/MIND/Ours
This editing result is shown in Table~\ref{tab:cog_main}, compared with non-meta-cognitive methods, our meta-cognitive editing method achieves SOTA results on our CogEdit benchmark.
\begin{table*}[htbp]
\centering
\caption{Ablation study of our method on \textbf{CogEdit}}
\vspace{-8px}
\resulttablefontsize
\resizebox{2\columnwidth}{!}{
\begin{tabular}{c c c | c c | c c }
\toprule
\multicolumn{1}{c}{Method} & \multicolumn{2}{c}{\textsc{Counterfactual}} & \multicolumn{2}{c}{\textsc{Boundary Editing}} & \multicolumn{2}{c}{\textsc{Noisy Editing}} \\
\cmidrule(r){2-3} \cmidrule(r){4-5} \cmidrule(r){6-7}
        & \multicolumn{1}{c}{Fidelity$\uparrow$} 
        & \multicolumn{1}{c}{Adaptability$\uparrow$} 
        & \multicolumn{1}{c}{Reliability$\uparrow$} 
        & \multicolumn{1}{c}{Compliance$\uparrow$} 
        & \multicolumn{1}{c}{Clarity@2$\uparrow$} 
        & \multicolumn{1}{c}{Clarity@4$\uparrow$}  \\ 
\hline
    \textit{only} MSV monitor module & 76.39 & 39.21 & 79.52 & 44.73 & 31.72 & 26.90 \\
    \textit{only} inner meta-mem & 97.91 & 47.31 & 97.89 & 42.69 & 54.46 & 52.27 \\
    Meta-mem + MSV module & 91.15 & 50.28 & 93.44 & 49.77 & 56.27 & 50.88 \\
    \textit{only} label refiner & 80.99 & 48.69 & 93.66 & 58.71 & 36.12 & 33.63 \\
\hline
    MIND(Ours) & \textbf{99.87} & \textbf{56.47} & \textbf{99.33} & \textbf{59.08} & \textbf{60.86} & \textbf{57.40} \\
\bottomrule
\vspace{-18px}
\end{tabular}
}
\label{tab:ablation_components}
\end{table*}

% note
\paragraph{Limitations of Cognitive-Based Editing Methods.} %小标题别删
% 给一个总体结论，他们做不了meta-cognitive editing，现在的差不多，然后从表格分析具体结论：
% 这里的结论的目的不是让你分析他们的结果怎么样，以及什么导致了他们这样的结果，这样太浅了，而是你从这个结果回归到任务的设计理念再到底下的本质原因，也就是他们是cognitive，不具备哪个meta-cognitive的子能力。
% (1) COUNTERFACTUAL下什么指标不行，结合定义来说明本质，也就是被反事实的条件影响之后，去掉这个条件就做不好了。然后点回self-awareness。
% (2) BOUNDARY EDITING下指标有什么现象，结合任务定义说明了他们对这个任务是怎么处理的（cognitive），然后点回monitoring
% （3）。。。

% Existing knowledge editing methods primarily focus on ensuring that the model produces the correct output in edited scenarios. However, these methods fail to facilitate the model's understanding of why a particular output is expected in a given context. Which means the lack of meta-cognitive ability. As shown in Table \ref{tab:cog_main} 我们发现
% （1）counterfactual场景下，T-Patcher、IKE、MEND的Adaptability普遍较低，说明这几种方法在虽然可能能学到反事实场景的知识，但是编辑的新知识会影响到去掉反事实条件后旧知识的应用，即没有self-aware of知晓何时应该丢弃旧知识的能力
% （2）Boundary场景下，cognitive-based 方法尽管在Reliability上可能表现较好，但是Compliance普遍较低，即表明这些方法证明这些方法无法知晓被编辑的新知识何时被应用，即缺乏boundary monitoring的能力。
% （3）Noisy场景下，T-Patcher很难平衡Reliability和Clarity，而IKE、WISE的Reliability很高，证明模型学到了含有噪声的知识，这是应当被避免的。同时IKE、MEND的Clarity较低，证明模型不知应该如何获取有用的部分，即缺少reflective thinking的能力。

Existing knowledge editing methods primarily focus on ensuring that the model produces the correct output in edited scenarios. However, these methods fail to enhance the model's understanding of why a particular output is expected in a given context, \textbf{indicating a lack of meta-cognitive ability}. As shown in Table \ref{tab:cog_main}, we observe the following: \textbf{(1) In counterfactual scenarios}, T-Patcher, IKE, and MEND exhibit generally low Adaptability, suggesting that although these methods may acquire knowledge in counterfactual settings, the newly edited knowledge interferes with the application of prior knowledge once the counterfactual condition is removed. This indicates a lack of self-awareness in recognizing when outdated knowledge should be discarded. \textbf{(2) In boundary scenarios}, while cognitive-based methods may achieve relatively high Reliability, they generally perform poorly in terms of Compliance. This suggests that these methods fail to determine when the newly edited knowledge should be applied, highlighting a deficiency in boundary monitoring. \textbf{(3) In noisy scenarios}, although T-Patcher, IKE, and WISE are able to perform edits on examples containing noise. However, their lower \textbf{Clarity} scores suggest that these methods struggle to isolate informative signals from noisy labels, reflecting a deficiency in reflective reasoning capabilities.

\paragraph{Advantages of MIND.} %这个就改成你的方法
% 这个时候(1) (2) (3)就不要先写cognitive method的结论了，你就围绕着meta-cognitive出发，体现你的优势点，然后哪块对比反映了你meta-cognitive的某个子能力。比如：ost cognitive-based419
% methods perform poorly when additional constraints are in-420
% troduced. This includes methods with high Reliability, such421
% as SERAC. 
% 你这里的写法显然是在说现有方法的能力不行，而不是说你的方法行
% 总之，这里的格式就是：短语加粗(更强的XXX能力)+在某个子任务下，MIND相比baseline在指标结果上怎么样(充篇幅可以来个举例)，说明了在这个任务下能怎么样怎么样（结合定义，有点像前面反过来），说明它具备了怎么样的能力

% Since the cognitive-based editing module cannot effectively extract useful information from editing samples, MIND demonstrates significantly improved results under such conditions.
% (1) 更强的self-aware能力使得MIND能在counterfactual场景下的编辑取得更高的Adaptability，说明MIND能够更清晰地认知到合适应该丢弃掉outdated知识
% （2）更强的boundary monitor能力使得MIND能够在boundary限制的场景下取得更高的Compliance，说明MIND能够更清晰地得知合适应该使用新获得的知识
% （3）更强的reflective thinking能力使得MIND能够在Noisy editing场景下取得更高的Clarity，说明MIND能够思考并获取到标签中有用的信息

Since the cognitive-based editing module struggles to effectively extract useful information from editing samples, MIND demonstrates significantly improved performance under such conditions. \textbf{(1) Enhanced self-awareness:} In counterfactual scenarios, MIND achieves higher Adaptability than cognitive-based methods, indicating its ability to better recognize when to discard outdated knowledge, demonstrating its superior capability of self-awareness. \textbf{(2) Stronger boundary monitoring:} In boundary-constrained scenarios, MIND attains higher Compliance than cognitive-based methods, showcasing its capability to accurately determine when to apply newly acquired knowledge, demonstrating its boundary monitoring ability. \textbf{(3) Improved reflective thinking:} In noisy editing scenarios, MIND achieves higher Clarity than cognitive-based methods, highlighting its ability to analyze and extract useful information from noisy labels, demonstrating its reflective thinking ability.

\subsection{Results on the MMEdit Dataset}  
\label{subsec:5.3}
In addition to evaluating meta-cognitive editing, we assess our method on the widely used cognitive-based editing dataset, MMEdit~\cite{DBLP:conf/emnlp/0008TL0WC023}, as shown in Table~\ref{tab:mmedit}. Compared to the reported multimodal knowledge editing results, our method also achieves competitive performance. We observe that: 
\textbf{(1)} Most of the cognitive-based methods can not effectively balance Reliability, Generality, and Locality.
\textbf{(2)} While cognitive-based methods such as MEND and WISE achieve high Reliability, they often do so at the expense of Generality, leading to a reduction in the generalizability of the edited knowledge.
\textbf{(3)} For the Locality metric, methods such as In-Context Editing and SERAC tend to perform suboptimally.
In contrast, MIND controls the activation of the meta-knowledge unit to regulate when both prior and newly acquired knowledge should be activated, thereby achieving a balance across multiple metrics. This ensures effective meta-cognitive editing while preserving cognitive ability.

\subsection{In Depth Analysis}
\label{subsec:5.5}

% \subsubsection{Benchmark Data Info}
% \paragraph{Effect of Meta-Memory}
% We test our meta-memory with different 

\paragraph{Effect of Individual Components.}
% 1) only inner meta-mem
% 2) only shapley competition module
% 3) meta-mem + competition module
% 4) only label enhancement
% % ----
% 5) Ours

% 我们测试了不同的component组合的情况下，我们方法的有效性：

% (1) only MSV monitor module: 我们freezed了meta-memory的参数为初始值，仅靠MSV module的约束下，meta-konwledge中几乎不存在知识特征，导致了几个场景下的指标都相对较低。

% (2) 仅有meta-memory: 在引入meta-memory之后，模型能够对FFN层的激活值重新分布，能够学习到编辑样例相关的知识，但是无法知道何时该使用新的知识与抑制旧知识，因此在Adaptability和Compliance都表现较差

% (3) meta-memory+MSV monitor module: 在MSV monitor module的更新指导下，模型能够高效地更新parameter of meta-memory，但是由于较难提取出label中的有用特征，因此clarity较低。

% (4) only label enhancement module: 我们从原有的模块中去掉了Meta-Memory和相关模块，结果展示出如果无法对FFN层的激活值重新分布，那么仅靠编辑时对label特征的加强很难控制神经元的激活通路，导致在counterfactual和boundary场景下的成功率有较为明显的下降。
\begin{figure}[tbp]
    \vspace{-8px}
    \centering
    \includegraphics[width=0.44\textwidth]{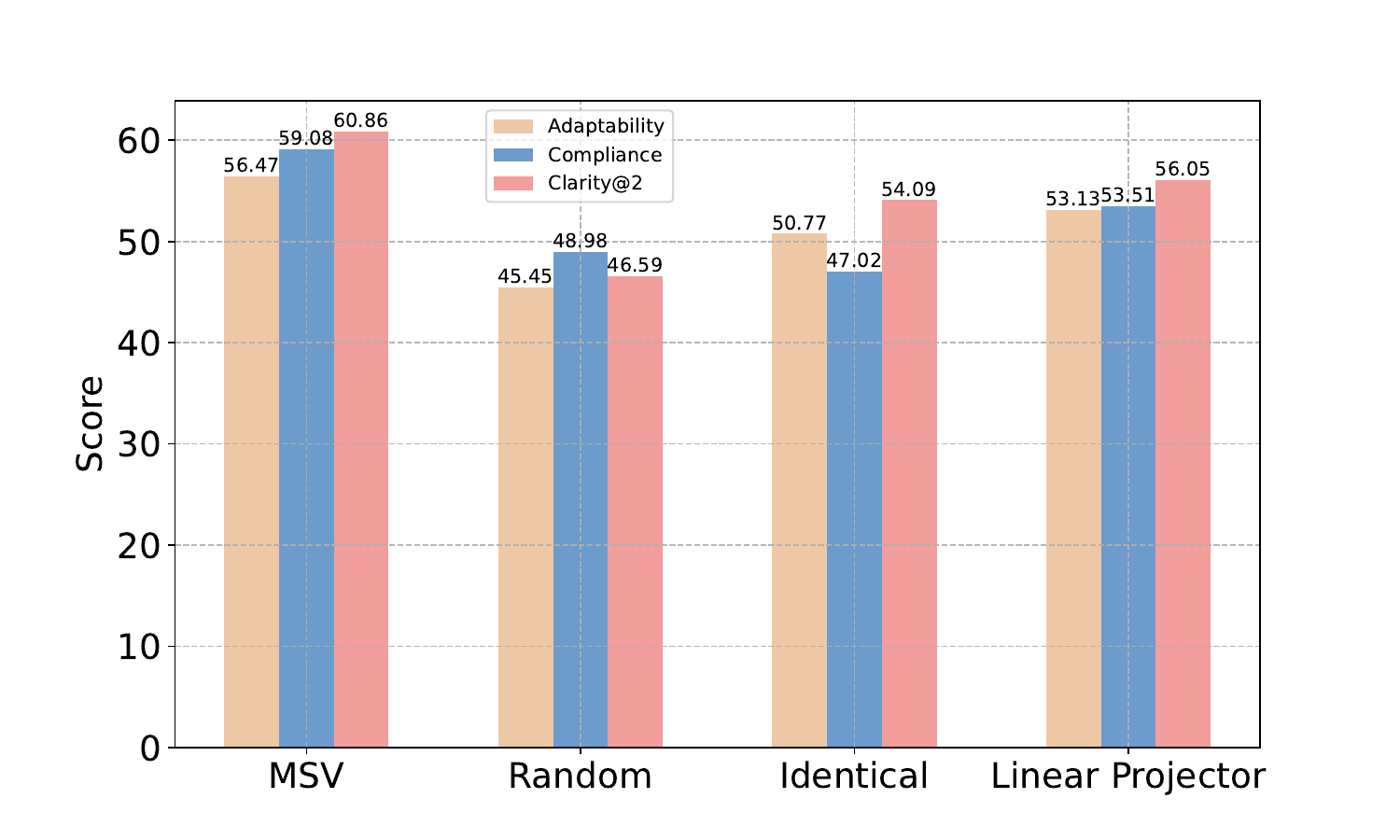}
    \vspace{-1.0em}
    \caption{Effectiveness of MSV-Monitored Update}
    \label{fig:ablation_shapley_bar}
    \vspace{-1.6em}
\end{figure}
We evaluate the effectiveness of our approach under different combinations of components. \textbf{(1) only MSV monitor module}, we froze the parameters of the meta-memory at their initial values, relying solely on the constraints imposed by the MSV module. As a result, the meta-knowledge contained minimal knowledge-related features, leading to relatively low performance across several scenarios. \textbf{(2) using only the meta-memory}, the model was able to redistribute the activation values of the FFN layers, allowing it to learn knowledge relevant to the editing samples. However, it was unable to determine when to utilize new knowledge or suppress outdated information, resulting in poor performance in terms of adaptability and compliance. \textbf{(3) meta-memory and the MSV monitor module}, the model effectively updated the parameters of the meta-memory under the guidance of the MSV monitor module. However, due to difficulties in extracting useful features from the labels, the clarity of the learned knowledge was relatively low. \textbf{(4) only label refiner}, we removed the meta-memory and its associated components. The results indicate that, without the ability to redistribute the activation values of the FFN layers, merely enhancing the label features during editing was insufficient to control neural activation pathways. This limitation led to a noticeable decline in success rates under counterfactual and boundary scenarios.

\paragraph{Effect of MSV monitor module.}
We replace our MSV monitor module by \textbf{(1) Random distribution}; \textbf{(2) Identical distribution:} means each meta-knowledge unit owns the same significance; \textbf{(3) Linear Projector:} as the simplified version of our module. Various modules are employed to monitor meta-memory updates and assess the effectiveness.
Figure~\ref{fig:ablation_shapley_bar} demonstrates that our meta-memory MSV monitor module outperforms other approaches, indicating its effectiveness in monitoring meta-memory and guiding the learning process.

% 表格/柱状图
% Identical Distribution
% Randomized
% Linear Projector
% Shapley-based (Ours)

% \subsubsection{Visualization of Reconstructed Knowledge Path}
% % reconstructed knowledge path分布 用图像证明，热力图

% To 更清楚地展示our prototype-based Meta-Label Refiner的信息提取能力，我们对比了添加与不添加encoder module情况下hidden state的分布情况，我们采用tsne降维处理以可视化分布。图\ref{fig:encoder_scatter}展示了在不添加Meta-Label Refiner的情况下，模型从noisy label信息中较难获取和当前场景有关的信息，但在添加encoder之后，模型更容易提取出标签中和当前语境相关的有用信息

\paragraph{Effect of the Meta-Label Refiner.}

\begin{figure}[tbp]
    \centering
    \includegraphics[width=0.42\textwidth]{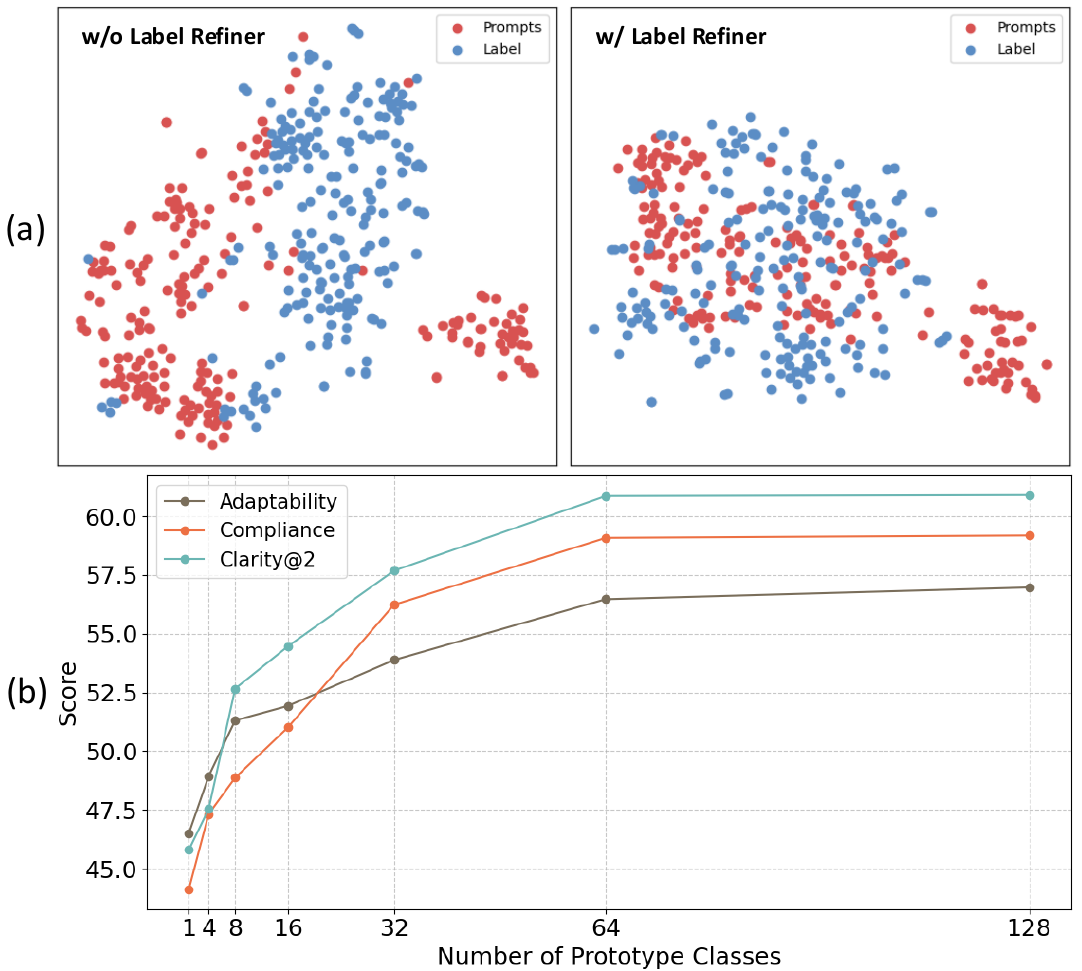}
    \vspace{-1em}
    \caption{Ablation study on meta-label refiner: (a) Alignment Visualization (b) Effect of Prototype Classes}
    \vspace{-1.6em}
    \label{fig:ablation_study}
\end{figure}

To better illustrate the information extraction capability of our prototype-based meta-label refiner, we compare the distribution of hidden states with and without the encoder module. We apply t-SNE for dimensionality reduction to visualize the distributions. ~\autoref{fig:ablation_study}(a) shows that, without meta-label refiner, the model struggles to extract relevant information from noisy labels, causing label-prompt misalignment. In contrast, the encoder improves alignment and prediction accuracy.

% However, after incorporating the encoder, the model effectively identifies and retains useful information that aligns with the current context.

% \begin{figure}[htbp]
%     \centering
%     \includegraphics[width=0.48\textwidth]{resources/i_ablation_hs.pdf}
%     \caption{Meta-Label Refiner Visualize}
%     \label{fig:ablation_encoder_scatter}
% \end{figure}

% \begin{figure}[htbp]
%     \centering
%     \includegraphics[width=0.48\textwidth]{resources/i_ablation_prototype.pdf}
%     \caption{Performance with Different Number of Prototype Classes}
%     \label{fig:ablation_prototype}
% \end{figure}

% \begin{figure}[tbp]
%     \centering
%     \begin{subfigure}{0.43\textwidth}
%         \centering
%         \includegraphics[width=\linewidth]{resources/i_ablation_hs.pdf}
%         % \vspace{0.1px}
%         \caption{Alignment Visualization}
%         \label{fig:ablation_encoder_scatter}
%     \end{subfigure}
%     \hfill
%     \begin{subfigure}{0.45\textwidth}
%         \centering
%         \includegraphics[width=\linewidth]{resources/i_ablation_prototype.pdf}
%         \caption{Effect of Prototype Classes}
%         \label{fig:ablation_prototype}
%     \end{subfigure}
%     \vspace{-1em}
%     \caption{Ablation study on meta-label refiner.}
%     \label{fig:ablation_study}
%     \vspace{-1em}
% \end{figure}

% We also test prototype class的数量对模型性能造成的影响，实验结果如图~\ref{fig:ablation_prototype}所示。更多种类的prototype意味着能引入更精确分类的contextual 信息特征，但是同时会带来较大的计算量消耗，而当prototype种类较少时，用于shift的contextual编码信息可能不够准确，因此对编辑过程的贡献较小。而当prototype种类较大时(例如K=128)，增大class数量对结果的贡献微乎其微，甚至可能产生副作用，因此在实验中，我们选用K=64作为setting。
We also investigate the impact of the number of prototype classes on the model's performance, as shown in ~\autoref{fig:ablation_study}(b). A greater variety of prototypes allows for the inclusion of more precise contextual information features for classification; however, this comes at the cost of increased computational overhead. On the other hand, when the number of prototypes is small, the contextual encoding information used for shifting may be less accurate, thus contributing less to the editing process. When the number of prototypes is large (e.g., \( K = 128 \)), increasing the number of classes has a negligible impact on the results and may even lead to computational cost. Therefore, we select \( K = 64 \) as the setting for the balance of computation cost and accuracy.

% \paragraph{Computation Cost Comparison}

% 折线图 三条线的变化趋势，以MiniGPT4/llava为例
% \paragraph{Editing in different layers}
% % 这章就是localization in WISE
% Research~\cite{geva2020transformer} have shown that deeper layers tend to encode more semantic information, and thus, editing or manipulating a model's behavior to reflect semantic changes should primarily focus on these layers. In contrast, shallower layers mainly capture surface-level patterns, such as repeated n-grams or syntactic features, which makes them more suitable for tasks involving token-level or structural modifications. 

% \paragraph{Qualitative examples}

\section{Conclusion}
% 在这篇文章中，我们基于现有的cognitive-based知识编辑数据集的不足，提出了CogEdit，一个用于衡量多模态知识编辑是否具有meta-cognitive能力的数据集，并提出了一个meta-cognitive的知识编辑方法MIND。我们发现现有的知识编辑方法fail to拥有meta cognition ability,并在traditional的editing场景下也能取得很好的效果， 在多个数据集上的实验证明了我们的结论

In this paper, we identify the limitations of existing cognitive-based knowledge editing datasets and propose \textbf{CogEdit}, a benchmark designed to assess the meta-cognitive capabilities of multimodal knowledge editing. 
We find that though current knowledge editing methods perform well in traditional editing scenarios, they fail to exhibit meta-cognitive abilities.
On this basis, we introduce \textbf{MIND}, a novel meta-cognitive knowledge editing approach. Extensive Experimental results demonstrate the effectiveness of MIND, which achieves superior performance on both cognitive and meta-cognitive multimodal editing benchmarks.

\clearpage
\section{Limitations}
Despite the promising results demonstrated in our study, several limitations remain that warrant further investigation.
First, many existing generative multimodal models~\cite{podell2023sdxlimprovinglatentdiffusion,ramesh2021zero,pan2025unlocking,ramesh2022hierarchical,pan2025generative} excel in tasks like image captioning and visual question answering. In this work, we focus solely on assessing the model's understanding of edited multimodal knowledge, without considering the impact of such edits on its generative abilities. Future work should explore how these edits affect the quality and consistency of multimodal generation.
Second, due to computational constraints, we were unable to conduct experiments on large-scale models, such as the 65B LLaMA Adapter V2~\cite{gao2023llamaadapterv2}. Scaling our method to such models poses additional challenges and is left for future exploration.

% Bibliography entries for the entire Anthology, followed by custom entries
% \bibliography{anthology,custom}
% Custom bibliography entries only
\bibliography{main}

\clearpage
\appendix
\label{sec:appendix}
\section{Dataset Details}
\label{appendix:dataset_details}
% To ensure diversity in data sources, we curated a total of 3,178 VQA questions from various multimodal QA datasets\cite{marino2019okvqavisualquestionanswering,hudson2019gqanewdatasetrealworld}. These questions span various specialized domains, including chart understanding, scientific knowledge, celebrity recognition, and emotion recognition. The selection process also considered different capability tiers, categorized from simplest to most complex: Perception, Recognition, Understanding, and Reasoning.
\subsection{Capability tiers in CogEdit Benchmark}
Data in CogEdit are considered to own different capability tiers, from low to high, the number of instances are shown in ~\autoref{tab:capability_tiers}

\begin{itemize}
    \item \textbf{Perception:} Scene understanding, instance identity, attribute recognition, and spatial localization.  
    \item \textbf{Multimodal Recognition:} Text recognition, object recognition.
    \item \textbf{Multimodal Understanding:} Visual reasoning and visual referring expressions.  
    \item \textbf{Cascading Reasoning:} Logical reasoning, numerical reasoning, and visual mathematics.  
\end{itemize}
To ensure the reliability and accuracy of the curated data, we conducted a two-pass quality control process. For the first pass, we utilized open-source models~\cite{chen2024internvlscalingvisionfoundation, grattafiori2024llama3herdmodels, li2024llava, wang2024qwen2} to automatically validate and refine the data. In the second pass, a manual review was performed to further ensure correctness in order to remove ambiguous or poorly phrased entries. This combination of automated and manual methods ensured consistency with the intended domain and capability tiers.

% \begin{table}[h]
%     \centering
%     \resizebox{\columnwidth}{!}{%
%     \begin{tabular}{|c|c|c|c|c|}
%         \hline
%         Capability & Perception & Recognition & Understanding & Reasoning \\
%         \hline
%         Number & 530 & 285 & 291 & 68 \\
%         \hline
%     \end{tabular}
%     }
%     \caption{Distribution of Instances Across Capability Tiers}
%     \label{tab:capability_tiers}
%     % \vspace{-2em}
% \end{table}

% \begin{table}[h]
%     \centering
%     \resizebox{\columnwidth}{!}{%
%     \begin{tabular}{c|c|c|c|c} % ❗去掉了最前和最后的 |
%         \hline
%         Capability & Perception & Recognition & Understanding & Reasoning \\
%         % \hline
%         Number & 530 & 285 & 291 & 68 \\
%         \hline
%     \end{tabular}
%     }
%     \caption{Distribution of Instances Across Capability Tiers}
%     \label{tab:capability_tiers}
%     % \vspace{-2em}
% \end{table}

\begin{table}[h]
    \centering
    \scriptsize
    \begin{tabularx}{\columnwidth}{*{5}{>{\centering\arraybackslash}X}}
        \toprule
        \textbf{Capability} & Perception & Recognition & Understanding & Reasoning \\
        \midrule
        \textbf{Number} & 530 & 285 & 291 & 68 \\
        \bottomrule
    \end{tabularx}
    % \vspace{-2em}
    \caption{Distribution of Instances Across Capability Tiers}
    \vspace{-1.5em}
    \label{tab:capability_tiers}
\end{table}

\subsection{Formal Task Definitions and Metrics in CogEdit}
% ========================
% generate \(\mathcal{C}(p_d) = (i_d, x_d) + \delta_{cf}\), where \(\delta_{cf}\) represents counterfactual modifications introduced either into the text or the image, respectively, ensuring that the modified VQA instance contradicts the factual assumptions or the original scene.

% \( p_{cf} = (i_{cf}, x_{cf}) \), where \( p_{cf} \) is defined as either \( p_{cf} = (i_d, x_d + x_{cf}) \) or \( p_{cf} = (i_d + i_{cf}, x_d) \). Here, \( x_{cf} \) and \( i_{cf} \) represent counterfactual modifications introduced into the text and image, respectively, ensuring that the modified VQA instance contradicts the factual assumptions or the original scene.

\paragraph{Level 1: Counterfactual condition-driven editing.}
% For counterfactual condition-driven editing, we evaluate the editing result with two metrics: \textbf{Fidelity}, which measures the success rate of editing in counterfactual scenarios, and \textbf{Adaptability}, which evaluates the model's success rate in answering factual questions after the counterfactual edit.

% \paragraph{Fidelity}  
% Editing fidelity refers to the ability to change the prediction from \( t_o \) to \( t_{cf} \). Therefore, our goal is to update the model's original parameters \( \theta_o \) to \( \theta_{cf} \), such that \( f(p_{cf}; \theta_{cf}) = t_{cf} \). The success rate of editing in a counterfactual scenario is described as:  
% \begin{equation}
%     \mathcal{M}_{fidelity} = \mathbb{E}_{(p_{cf}) \sim \mathcal{C}(p_d) } \left[ \mathds{1}_{f \left( p_{cf}; \theta_{cf} \left( p_{cf}, t_{cf} \right) \right) = t_{cf}} \right]
% \end{equation}

% \paragraph{Adaptability}  
% Editing adaptability is used to test whether the model has understood the old knowledge and can clearly identify the scenarios in which the old knowledge should no longer be applied. Therefore, we use \( p_d \) to evaluate the editing outcome. Adaptability is described by the following:

% \begin{equation}
% \mathcal{M}_{adaptability} = \mathbb{E}_{(p_d) \sim \mathcal{D}_{\textrm{base}}} \left[ \mathds{1}_{f \left( p_{d}; \theta_{cf} \right) = t_d} \right]
% \end{equation}
% where \( \theta_{cf} \) refers to the edited parameters.

For counterfactual condition-driven editing, we evaluate the editing result with two metrics: \textbf{Fidelity}, which measures the success rate of editing in counterfactual scenarios, and \textbf{Adaptability}, which evaluates the model's success rate in answering factual questions after the counterfactual edit.

Formally, for a given instance $(i_e, x_e)$ (\textit{e.g.}, $i_e$ is an image of ``\textit{Solar System}''; $x_e$=``\textit{Which is the smallest planet in this galaxy?}'') with original correct answer $y$=``\textit{Mercury}'', 
we introduce a counterfactual assumption $cf_e$=``\textit{Assume that Pluto is a planet, rather than a dwarf planet}'', which adjusts the correct answer to $y_e$=``\textit{Pluto}'' with the editing objective as $\mathds{1}_{f(cf_e, i_e, x_e; \theta_e(cf_e, i_e, x_e)) = y_e}$\footnote{$\mathds{1}_{\cdot}$ denotes the indicator function: it returns 1 if the condition is true, 0 otherwise.} (we use \textbf{Fidelity} as the editing success rate).
After editing, we remove the counterfactual assumption with $y_o$ restored to the correct answer, assessing whether the edited model can recall its prior correct knowledge when the assumption no longer holds: $\mathds{1}_{f(i_e, x_e; \theta_e(cf_e, i_e, x_e))=y}$ (where \textbf{Adaptability} quantifies the success rate of answering correctly after removing the counterfactual assumption).

\paragraph{Fidelity.}  
Editing fidelity refers to the ability to change the prediction to \( y_e \). Therefore, our goal is to update the model's original parameters \( \theta_o \) to \( \theta_e \), such that \( f(cf_e, i_e, x_e; \theta_e) = y_e \). The success rate of editing in a counterfactual scenario is described as:

% \begin{equation}
%     \mathcal{M}_{fidelity} = \mathbb{E}_{(cf_e, i_e, x_e) \sim \mathcal{C}(i_e, x_e) } \left[ \mathds{1}_{f \left( cf_e, i_e, x_e; \theta_e \left( cf_e, i_e, x_e \right) \right) = y_e} \right]
% \end{equation}

\begin{align}
\mathcal{M}_{fidelity} 
&= \mathbb{E}_{(cf_e, i_e, x_e) \sim \mathcal{C}(i_e, x_e)} \big[ \nonumber \\
&\quad \mathds{1}_{f \left( cf_e, i_e, x_e;\; \theta_e \left( cf_e, i_e, x_e \right) \right) = y_e} \big]
\end{align}

where \(\mathcal{C}(i_e, x_e)\) denotes the distribution of counterfactual instances derived from \( (i_e, x_e) \), $y_e$ is the correct answer of the question $(cf_e, i_e, x_e)$ with counterfactual assumptions.

\paragraph{Adaptability.}  
Editing adaptability is used to test whether the model has retained its prior correct knowledge and can revert to the original prediction when the counterfactual assumption is removed. Adaptability is described by the following:

% \begin{equation}
% \mathcal{M}_{adaptability} = \mathbb{E}_{(i_e, x_e) \sim \mathcal{D}_{\textrm{base}}} \left[ \mathds{1}_{f \left( i_e, x_e; \theta_e \left( cf_e, i_e, x_e \right)\right) = y} \right]
% \end{equation}

\begin{align}
\mathcal{M}_{adaptability} 
&= \mathbb{E}_{(i_e, x_e) \sim \mathcal{D}_{\textrm{base}}} \big[ \nonumber \\
&\quad \mathds{1}_{f \left( i_e, x_e;\; \theta_e \left( cf_e, i_e, x_e \right) \right) = y} \big]
\end{align}

where \( \theta_e \) refers to the edited parameters, $y$ is the correct answer of the question $(i_e, x_e)$.

\paragraph{Level 2: Boundary-restricted Editing.}

% \paragraph{Reliability}
% Reliability in this context refers to the model's ability to maintain correct predictions based on the original knowledge when applied in the original scenario. It quantifies the model's success rate in providing accurate responses using the original parameters \( \theta_b \). The reliability metric can be defined as follows:

% \begin{equation}
%     \mathcal{M}_{reliability} = \mathbb{E}_{(p_d) \sim \mathcal{D}_{\textrm{base}}} \left[ \mathds{1}_{f \left( p_d; \theta_d \left( p_d, t_d \right) \right) = t_d} \right]
% \end{equation}
% where \( \theta_b \) refers to the parameters associated with the original scenario, \( p_d \) is the base VQA sample, and \( t_d \) is the correct answer for that sample.

% \paragraph{Compliance}  
% Boundary Compliance measures the model's success in correctly identifying when the new knowledge should be applied, ensuring that the modified response aligns with the constraints of the new scenario. This metric evaluates the model's ability to provide accurate answers when the original knowledge is no longer applicable. The Boundary Compliance metric is defined as:
% \begin{equation}
%     \mathcal{M}_{compliance} = \mathbb{E}_{(p_b) \sim \mathcal{C}(p_d)} \left[ \mathds{1}_{f \left( p_b; \theta_d \right) = t_b} \right]
% \end{equation}
For boundary-restricted editing, we evaluate the editing performance using two metrics: \textbf{Reliability}, which measures the success rate of editing under the original, unaltered context, and \textbf{Compliance}, which assesses the model’s ability to appropriately constrain the activation of edited knowledge when additional boundary conditions are introduced. Reliability reflects whether the model has correctly internalized the intended factual update, while Compliance tests whether the model can generalize this knowledge appropriately without overgeneralization or misuse in contextually incompatible settings.

Formally, for the previously given instance \( (i_e, x_e) \) with editing target $y_e$=``\textit{Mercury}'', we conduct the conventional editing process to let the model remember the correct answer in the original context, with the editing objective as $\mathds{1}_{f(i_e, x_e; \theta_e(i_e, x_e)) = y_e}$ (we use \textbf{Reliability} as the editing success rate).
After editing, we construct a new problem instance by introducing boundary constraints  $b_e$=``\textit{Considering the five farthest planets from the central star}'', which limits the available candidate planets and adjusts the correct answer to $y'_e$=``\textit{Mars}'', in order to assess whether the edited model knows the boundary to utilize the edited knowledge rather than  indiscriminately activating it: $\mathds{1}_{f(b_e, i_e, x_e; \theta_e(i_e, x_e))=y'_e}$ (where \textbf{Compliance} quantifies the success rate of answering correctly after adding the boundary constraint assumption).

\paragraph{Reliability}
Reliability in this context refers to the model's ability to edit successfully based on the original knowledge when applied in the original scenario. It quantifies the model's success rate in providing accurate responses using the edited parameters \( \theta_e \). The reliability metric can be defined as follows:

\begin{equation}
    \mathcal{M}_{reliability} = \mathbb{E}_{(i_e, x_e) \sim \mathcal{D}_{\textrm{base}}} \left[ \mathds{1}_{f \left( i_e, x_e; \theta_e(i_e, x_e) \right) = y_e} \right]
\end{equation}
% where \( \theta_e(i_e, x_e) \) refers to the parameters after editing, \( (i_e, x_e) \) is the base VQA sample, and \( y_o \) is the correct answer for that sample.
Where $y_e$ is the correct answer of the question $(i_e, x_e)$.

\paragraph{Compliance}  
Compliance measures the model's success in correctly identifying when the new knowledge should be applied, ensuring that the modified response aligns with the constraints of the new scenario. This metric evaluates the model's ability to know when the new knowledge is applicable. The Compliance metric is defined as:
% \begin{equation}
%     \mathcal{M}_{compliance} = \mathbb{E}_{(b_e, i_e, x_e) \sim \mathcal{C}(i_e, x_e)} \left[ \mathds{1}_{f \left( b_e, i_e, x_e; \theta_e(i_e, x_e) \right) = y'_e} \right]
% \end{equation}

\begin{align}
\mathcal{M}_{compliance}
&= \mathbb{E}_{(b_e, i_e, x_e) \sim \mathcal{C}(i_e, x_e)} \big[ \nonumber \\
&\quad \mathds{1}_{f \left( b_e, i_e, x_e;\; \theta_e(i_e, x_e) \right) = y'_e} \big]
\end{align}

Where $y'_e$ is the correct answer of the boundary restricted question $(b_e, i_e, x_e)$.

\paragraph{Level 3: Noisy Label Editing.}

To systematically evaluate the model's adaptability to noisy inputs, we propose a new evaluation metric, Clarity@K, which quantifies the model's ability to maintain accuracy as noise increases.  

Formally, for a the same instance $(i_e, x_e)$ with originally correct answer $y$=``\textit{Mercury}'' , 
% $y_e$=$f(i_e, x_e; \theta)$, 
we introduce a perturbation $noise_e$=``\textit{Nebula, Cosmos.}'', which are some non-planet labels, to the originally correct prediction, which adjusts the correct answer to $y_e$ containing noise.
After editing, the noisy pertubation is removed to assess whether the edited model can get to know the useful part in edited knowledge: $\mathds{1}_{f(i_e, x_e; \theta_e(noise_e, i_e, x_e))=y}$( where \textbf{Clarity@K} quantifies the model's ability to maintain accuracy as noise increases and $K$ denotes the number of noisy elements).

% \paragraph{Robustness}  
% Robustness measures the model’s ability to correctly apply the intended edit under noisy conditions. A higher Robustness score indicates that the model consistently produces accurate outputs, even when faced with ambiguous or corrupted labels. The formal definition of Robustness is given by:  

% \begin{equation}
%     \mathcal{M}_{\text{robustness}} = \mathbb{E}_{(noise_e, i_e, x_e) \sim \mathcal{C}(i_e, x_e, k) } \left[ \mathds{1}_{f \left( i_e, x_e; \theta_e \left( noise_e, i_e, x_e \right) \right) = y_e} \right]
% \end{equation}  

% Where \(\mathcal{C}(i_e, x_e, k)\) denotes the function that generates noisy-labeled instances. $y_e$ is the editing target containing noisy labels.

\paragraph{Clarity@K}
Editing clarity is defined to quantify the editing success rate after editing with the presence of noisy labels. Higher values of Clarity@K indicate the model can clearly discern and edit the target knowledge despite the noise. The Clarity@K metric is defined as:  

% \begin{equation}
%     \mathcal{M}_{\text{Clarity@K}} = \mathbb{E}_{(i_e, x_e) \sim \mathcal{D}_{\textrm{base}} } \left[ \mathds{1}_{f \left( i_e, x_e; \theta_e \left( noise_e, i_e, x_e \right) \right) = y} \right]
% \end{equation}

\begin{align}
\mathcal{M}_{Clarity@K}
&= \mathbb{E}_{(i_e, x_e) \sim \mathcal{D}_{\textrm{base}}} \big[ \nonumber \\
&\quad \mathds{1}_{f \left( i_e, x_e;\; \theta_e \left( noise_e, i_e, x_e \right) \right) = y} \big]
\end{align}

Where $y$ is the correct answer of the question $(i_e, x_e)$.
% This evaluation highlights the model's ability to provide feedback and summarization throughout the editing process, showcasing its capacity to reflect on the provided information and discern the correct knowledge boundaries. By analyzing Clarity@K, we aim to gain insights into the effectiveness of the model in overcoming label ambiguity and maintaining editing precision.  

\begin{table*}[t]
    \centering
    \caption{Lifelong editing results of LLaVA on \textbf{MMEdit} for different values of $T$ ($T$: Number of Edits.)}
    \vspace{-8px}
    \resulttablefontsize
    \setlength{\tabcolsep}{4pt}
    \resizebox{2\columnwidth}{!}{
    \begin{tabular}{l c c c c c c | c c c c c }
    \toprule
    \multicolumn{2}{c}{} & \multicolumn{5}{c}{$T=10$}  & \multicolumn{5}{c}{$T=100$}  \\
    \cmidrule(r){3-7}  \cmidrule(r){8-12}
    & \multicolumn{1}{c}{Method} &  \multicolumn{1}{c}{Reliability}$\uparrow$  & \multicolumn{1}{c}{T-Generality}$\uparrow$  & \multicolumn{1}{c}{M-Generality}$\uparrow$  & \multicolumn{1}{c}{T-Locality }$\uparrow$ & \multicolumn{1}{c}{M-Locality}$\uparrow$ & \multicolumn{1}{c}{Reliability}$\uparrow$ & \multicolumn{1}{c}{T-Generality}$\uparrow$ & \multicolumn{1}{c}{M-Generality}$\uparrow$ & \multicolumn{1}{c}{T-Locality}$\uparrow$ & \multicolumn{1}{c}{M-Locality}$\uparrow$ \\
    \hline
    \multicolumn{12}{c}{\textsc{Editing VQA}} \\
    \hline
    \multirow{2}{*}{Base Methods} 
    % & Base Model &  &  &  &  &  &  &  &  &  &   \\
    & FT (vision block) & 49.26 & 46.01 & 42.13 & 100.00 & 30.42 & 41.65 & 42.53 & 30.41 & 98.92 & 26.48 \\
    & FT (last layer) & 53.84 & 50.29 & 45.03 & 91.61 & 63.45 & 39.09 & 34.06 & 38.07 & 92.84 & 57.04 \\

    \hline
    \multirow{4}{*}{Cognitive Editing} 
    & T-Patcher & 82.16 & 76.45 & 76.55 & 71.10 & 74.50 & 70.57 & 63.39 & 61.12 & 36.42 & 48.90  \\
    % & Knowledge Editor &  &  &  &  &  &  &  &  &  &   \\
    & SERAC & 88.09 & 83.40 & 83.57 & 64.91 & 15.50 & 88.08 & 81.53 & 82.48 & 62.13 & 12.90   \\
    & MEND & 22.58 & 22.30 & 21.48 & 32.40 & 18.22 & 11.43 & 19.66 & 19.31 & 8.95 & 9.37  \\
    & WISE & 55.04 & 52.71 & 70.99 & 100.0 & 92.17 & 27.53 & 27.54 & 21.63 & 100.0 & 83.48  \\
    \hline
    Meta-Cognitive Editing  & \textbf{MIND(Ours)} & 84.88 & 84.44 & 78.82 & 91.00 & 75.23 & 78.49 & 70.79 & 65.60 & 88.72 & 73.34   \\
    \hline
    \multicolumn{12}{c}{\textsc{Editing Image Caption}} \\
    \hline
    \multirow{2}{*}{Base Methods} 
    % & Base Model &  &  &  &  &  &  &  &  &  &   \\
    & FT (vision block) & 46.12 & 47.56 & 43.20 & 100.00 & 84.94 & 45.96 & 47.50 & 42.93 & 99.9 & 71.05  \\
    & FT (last layer) & 69.52 & 68.24 & 63.47 & 93.69 & 74.71 & 65.14 & 60.11 & 56.23 & 81.05 & 70.09 \\

    \hline
    \multirow{4}{*}{Cognitive Editing}
    & T-Patcher & 79.94 & 80.42 & 74.52 & 54.92 & 68.06 & 59.88 & 62.81 & 54.37 & 26.08 & 39.99  \\
    % & Knowledge Editor &  &  &  &  &  &  &  &  &  &   \\
    & SERAC & 56.57 & 56.88 & 52.62 & 59.96 & 14.66 & 53.35 & 53.70 & 49.41 & 48.04 & 17.25  \\
    & MEND & 66.49 & 64.77 & 58.72 & 87.25 & 86.15 & 56.83 & 56.89 & 53.07 & 87.63 & 84.64  \\
    & WISE & 73.10 & 71.21 & 64.79 & 88.46 & 91.80 & 60.56 & 57.10 & 54.16 & 95.61 & 93.96  \\
    \hline
    Meta-Cognitive Editing  & \textbf{MIND(Ours)} & 86.12 & 85.70 & 79.13 & 93.43 & 77.10 & 75.76 & 71.82 & 61.51 & 90.55 & 69.54 \\
    \bottomrule
    % \vspace{-1em}
    \end{tabular}}
    \label{tab:lifelong_t10_t100}  
\end{table*}

\section{Experimental Details}
\label{appendix:implementation_details}
\subsection{Cognitive Dataset Details}
\paragraph{MMEdit}
In the context of Multimodal Large Language Models (MLLMs), which have shown great promise in integrating and understanding both visual and textual information, MMEdit~\cite{DBLP:conf/emnlp/0008TL0WC023} has been selected as the representative of the cognitive-based multimodal knowledge editing benchmark, which comprises two subtasks: Editing VQA (E-VQA)~\cite{antol2015vqa} and Editing Image Captioning (E-IC), with data selected from VQAv2~\cite{goyal2017making} and COCO Caption~\cite{chen2015microsoft}, respectively.These tasks aim to assess a model's ability to update its understanding of visual and textual inputs while ensuring stability and generalization across different modalities. MMEdit introduces a suite of metrics to evaluate the reliability, locality, and generality of model edits in multimodal settings. Each edit sample includes five examples that measure the following metrics. Reliability measures the accuracy of editing target inputs, locality evaluates the model's stability on unrelated knowledge (both textual and multimodal), and generality assesses the model's ability to generalize to rephrased inputs (textual and visual).

\subsection{Baseline Methods}
\paragraph{Fine-tune} Fine-tuning, in this context, refers to standard fine-tuning, where the model is adapted to new knowledge without employing any advanced tuning techniques. This is a straightforward adaptation process, and while more sophisticated methods~\cite{li2021prefix, pan2023self, liu2021p} or prompt-based tuning could be explored, we focused on traditional fine-tuning as a baseline for comparison. In our experiments, we conducted fine-tuning on the \textbf{last layer} of the MLLM and the \textbf{vision block}(vision projector for LLaVA and Qformer for MiniGPT4 respectively) separately.

\paragraph{Transformer-Patcher} ~\cite{huang2023transformer} is an editing method that enables continuous correction of mistakes in Transformer-based language models. The method adds and trains a few neurons (patches) in the last Feed-Forward Network (FFN) layer to correct specific errors without affecting overall model performance.

\paragraph{MEND} (Model Editor Networks with Gradient Decomposition)~\cite{mitchell2021fast} is a scalable method for efficiently editing large pre-trained neural network models using a single input-output pair. By leveraging the low-rank structure of fine-tuning gradients, MEND trains lightweight auxiliary networks to transform these gradients into targeted parameter updates. This approach enables fast, local, and reliable edits to model behavior without significantly affecting unrelated inputs.

\paragraph{SERAC} (Semi-Parametric Editing with a Retrieval-Augmented Counterfactual Model) ~\cite{mitchell2022memory} is a knowledge editing method requiring external memory. SERAC stores edits in an external memory and uses a scope classifier and counterfactual model to determine when and how to apply these edits.

\paragraph{IKE}(In-context Knowledge Editing)~\cite{zheng2023can} explores the potential of in-context learning (ICL) for editing factual knowledge in large language models without modifying their parameters. IKE constructs context demonstrations to inject new facts, update existing knowledge, and retain unrelated facts. IKE draws on relevant methods and concepts from the field of knowledge retrieval~\cite{lewis2020retrieval, izacard2021unsupervised, pan2024i3}.

\paragraph{WISE} ~\cite{wang2024wiserethinkingknowledgememory} is designed for lifelong model editing that bridges the gap between long-term and working memory in large language models. WISE proposes a dual parametric memory scheme, comprising a main memory for pre-trained knowledge and a side memory for edited knowledge. WISE also incorporates a knowledge-sharding mechanism to manage multiple edits efficiently and merge them without conflicts. 

\subsection{Implementation Details}
For MIND, we adopt the Adam optimizer with a learning rate of 8e-4 to update all parameters during the editing stage. Each one-step edit takes approximately 6–8 seconds to complete. During the pre-training phase of the meta-label refiner, we extract 5K labeled data samples for supervision, and the entire pre-training process takes less than one hour. All experiments are conducted on a computing cluster equipped with five NVIDIA RTX A6000 GPUs.

\begin{algorithm*}
\caption{Algorithmic Overview of the MIND Editing Process}
\label{alg:mind}
\KwIn{Activation vector $q_{\text{input}} \in \mathbb{R}^{d'}$, Target edit label $y$}
\KwOut{Refined activation $q_{\text{refined}}$}

\KwInit{Meta-memory matrix $\texttt{Mem} = \{m_i\}_{i=1}^{d'}$ \\
\hspace{1.1cm} MSV Monitor Network $\phi(\cdot)$ \\
\hspace{1.1cm} Prototype bank $\mathcal{P} = \{\mathbf{m}_k\}_{k=1}^{K}$ \\
\hspace{1.1cm} Label projection matrix $\mathbf{W}_p$, balance factor $\beta \in [0,1]$}

\vspace{1mm}
\Comment{Step 1: Self-Aware Meta-Memory Activation}
\For{$i \leftarrow 1$ \KwTo $d'$}{
    $q_{\text{meta}}[i] \leftarrow m_i(q_{\text{input}}[i])$ \tcp*{Apply meta-memory transformation}
}

\vspace{1mm}
\Comment{Step 2: Game-Theoretic Monitoring via MSV}
$\phi \leftarrow \text{Softmax}(\text{MSV\_Monitor}(q_{\text{meta}}))$ \tcp*{Shapley-based relevance}
\For{$i \leftarrow 1$ \KwTo $d'$}{
    $q_r[i] \leftarrow (q_{\text{meta}}[i] \cdot m_i) \cdot \phi[i]$ \tcp*{Weighted activation}
}

\vspace{1mm}
\Comment{Step 3: Reflective Meta-Label Refinement}
$p \leftarrow \text{RetrievePrototype}(\mathcal{P}, y)$ \tcp*{Get label prototype}
$\Delta \leftarrow \mathbf{W}_p \cdot p$ \tcp*{Project to label shift space}
$q_{\text{refined}} \leftarrow (1 - \beta) \cdot q_r + \beta \cdot \Delta$ \tcp*{Final refined activation}
\Return $q_{\text{refined}}$
\end{algorithm*}

\section{Results on Lifelong Editing Scenario}
\label{appendix:lifelong}

Lifelong model editing~\cite{wang2024wiserethinkingknowledgememory, hartvigsen2023aging} aims to edit the specific model over and over again, which enables large language models to continuously update and adapt their knowledge over time without requiring full retraining.

We also evaluated the effectiveness of our method in a lifelong editing scenario. To ensure fairness and eliminate task-induced biases, experiments were conducted on MMEdit (excluding IKE, as lifelong editing is not meaningful for IKE), with results presented in ~\autoref{tab:lifelong_t10_t100}. Compared to cognitive-based methods, MIND also demonstrates strong performance. We observe that: \textbf{(1)} Most methods, such as T-Patcher and MEND, exhibit a significant decline in all metrics after multiple editing iterations. However, as \( T \) increases, the rate of decline slows. In contrast, MIND maintains relatively high editing performance even as \( T \) increases.  \textbf{(2)} Editing methods specifically designed for lifelong learning, such as WISE, achieve high Locality scores. However, they fail to maintain consistently high Reliability and Generality. MIND is able to sustain relatively superior performance compared to other methods, indicating that it can continuously learn meta-knowledge from ongoing edits while maintaining effectiveness.

\section{MIND Method Pseudocode Implementation}
The pseudocode implementation of the MIND method is presented in ~\autoref{alg:mind}, which appears on the following page due to space constraints.

\section{Ethical Impacts}
This work poses no ethical concerns. All experiments are conducted using publicly available datasets and models, without involving any private data or subjective evaluations.

\end{CJK*}
\end{document}